\documentclass[10pt]{article} 
\usepackage[accepted]{tmlr}

\usepackage{amsmath,amsfonts,bm}

\def\eqref#1{equation~\ref{#1}}

\def\1{\bm{1}}

\DeclareMathAlphabet{\mathsfit}{\encodingdefault}{\sfdefault}{m}{sl}
\SetMathAlphabet{\mathsfit}{bold}{\encodingdefault}{\sfdefault}{bx}{n}

\usepackage{hyperref}
\usepackage{url}
\usepackage{booktabs}       
\usepackage{subcaption}     
\usepackage{amsfonts}       
\IfFileExists{nicefrac.sty}{\usepackage{nicefrac}}{}       
\usepackage{microtype}      
\usepackage{xcolor}         
\usepackage{amssymb}
\IfFileExists{algorithm.sty}{\usepackage{algorithm}}{%
  \usepackage{float}
  \newfloat{algorithm}{t}{loa}
  \newfloat{algorithm*}{t}{loa}
}
\let\tmlrAND\AND
\let\AND\relax
\IfFileExists{algorithmic.sty}{\usepackage{algorithmic}}{%
  \newenvironment{algorithmic}[1][]{\begin{list}{}{\leftmargin=1.5em\itemsep=0pt}}{\end{list}}
  \newcommand{\STATE}{\item}
  \newcommand{\WHILE}[1]{\item \textbf{while} ##1 \textbf{do}}
  \newcommand{\ENDWHILE}{\item \textbf{end while}}
}
\usepackage{graphicx}
\IfFileExists{scalerel.sty}{\usepackage{scalerel}}{}
\IfFileExists{multirow.sty}{\usepackage{multirow}}{\newcommand{\multirow}[3]{##3}}
\IfFileExists{wrapfig.sty}{\usepackage{wrapfig}}{\newenvironment{wrapfigure}[2]{\begin{figure}}{\end{figure}}}
\usepackage{xspace}
\usepackage{tcolorbox}

\usepackage{amsmath}
\definecolor{darkblue}{rgb}{0, 0, 0.5}
\definecolor{DeepBlue}{RGB}{0, 0, 200}
\hypersetup{colorlinks=true, citecolor=darkblue, linkcolor=darkblue, urlcolor=darkblue}

\newcommand{\adarft}{\textsc{AdaRFT}\xspace}

\newcommand{\githubcode}{\href{https://github.com/limenlp/verl}{github.com/limenlp/verl}}
\newcommand{\huggingface}{\href{https://huggingface.co/datasets/lime-nlp/DeepScaleR_Difficulty}{huggingface.co/datasets/lime-nlp/DeepScaleR\_Difficulty}}

\title{Efficient Reinforcement Finetuning via Adaptive Curriculum Learning}

\author{\name Taiwei Shi \email taiweish@usc.edu \\
      \addr University of Southern California
      \tmlrAND
      \name Yiyang Wu \email wendywu2@andrew.cmu.edu \\
      \addr Carnegie Mellon University
      \tmlrAND
      \name Linxin Song \email linxinso@usc.edu \\
      \addr University of Southern California
      \tmlrAND
      \name Tianyi Zhou \email tianyi.zhou@mbzuai.ac.ae \\
      \addr Mohamed Bin Zayed University of Artificial Intelligence
      \tmlrAND
      \name Jieyu Zhao \email jieyuz@usc.edu \\
      \addr University of Southern California\\
}

\def\month{06}  
\def\year{2026} 

\begin{document}

\maketitle

\begin{abstract}
Reinforcement finetuning (RFT) has shown great potential for enhancing the mathematical reasoning capabilities of large language models (LLMs), but it is often sample- and compute-inefficient, requiring extensive training. In this work, we introduce \adarft (\textit{Adaptive Curriculum Reinforcement Finetuning}), a method that significantly improves the efficiency of RFT through adaptive curriculum learning. \adarft dynamically adjusts the difficulty of training problems based on the model’s recent reward signals, ensuring that the model consistently trains on tasks that are challenging but solvable. This adaptive sampling strategy accelerates learning by maintaining an optimal difficulty range, avoiding wasted computation on problems that are too easy or too hard. \adarft requires only a lightweight extension to standard RFT algorithms like Proximal Policy Optimization (PPO), without modifying the reward function or model architecture. Experiments on competition-level math datasets demonstrate that \adarft improves convergence efficiency and reasoning performance. Given problem-level difficulty annotations, \adarft reduces RFT training time by up to 2$\times$ across data distributions and model scales, offering a more scalable and effective RFT framework.
\end{abstract}

\section{Introduction}
Reinforcement Finetuning (RFT) has emerged as a powerful technique for aligning large language models (LLMs) with task-specific goals, particularly in domains such as mathematics and code generation where correctness is well defined \citep{deepseekai2025deepseekr1incentivizingreasoningcapability, openai2024openaio1card}. By optimizing a policy model with reward signals that reflect task success, RFT enables more targeted learning than supervised finetuning (SFT) alone. However, despite its promise, RFT remains sample-inefficient and computationally expensive. Its training involves repeated rollout generation, reward computation, and policy updates—making it costly and difficult to scale \citep{ahmadian2024basicsrevisitingreinforcestyle, kazemnejad2024vineppounlockingrlpotential, li2024remaxsimpleeffectiveefficient, hu2025reinforcesimpleefficientapproach, cui2025processreinforcementimplicitrewards}. Recent efforts to address RFT inefficiency have focused on algorithmic simplification (e.g., RAFT \citep{dong2023raftrewardrankedfinetuning}, GRPO \citep{deepseekai2025deepseekr1incentivizingreasoningcapability}, ReMax \citep{li2024remaxsimpleeffectiveefficient}), and data-centric strategies (e.g., LIMO \citep{ye2025limoreasoning}, LIMR \citep{li2025limrrlscaling}). While these approaches improve sample or compute efficiency, they often introduce trade-offs: algorithmic simplifications may increase variance or limit stability, and static data filtering or scoring can be brittle, computationally heavy, or model-specific. Moreover, most methods' success relies on fixed datasets or training schedules, which can be suboptimal in non-uniform or imbalanced data regimes. More recently, early efforts have introduced curriculum-like ideas into RFT. Staged curricula divide training into a few manually-defined phases of increasing difficulty \citep{wen2025lightr1curriculumsftdpo, deepscaler2025, song2025fastcurlcurriculumreinforcementlearning}, but these are coarse-grained and lack adaptivity. Other methods use online data filtering, repeatedly rolling out and pruning training samples until the model’s average reward meets a target threshold \citep{bae2025online, yu2025dapoopensourcellmreinforcement}. While this approach helps prevent the model from stagnating on problems that are either too easy or too difficult, it is not truly adaptive and incurs significant rollout overhead.

To address these limitations, we propose \adarft, a reinforcement finetuning method based on adaptive curriculum learning \citep{bengio2009curriculumlearning}, which dynamically adjusts training set difficulty to match the model's evolving skill level. The intuition is simple: learning is most effective when tasks are neither too easy nor too hard. \adarft formalizes this by maintaining a target difficulty level, which increases or decreases based on recent reward feedback. At each step, the model is trained on examples closest to this target, promoting a steady progression through solvable yet challenging tasks. The full algorithm is outlined in Algorithm \ref{alg:curriculum}. Unlike prior work that relies on fixed stages, repeated rollouts, or model-specific data processing, \adarft is lightweight, general, and model-agnostic. It can be directly applied on top of any standard reinforcement learning (RL) algorithms like Proximal Policy Optimization (PPO) \citep{schulman2017proximalpolicyoptimizationalgorithms}.  We evaluate \adarft on a dataset spanning a wide range of competition-level math problems, including AMC, AIME, and IMO-style questions. Across multiple training distributions and two model sizes, \adarft improves training efficiency while maintaining or often improving final performance in our runs. Gains are especially notable in imbalanced data regimes, where static sampling often fails. Given problem-level difficulty annotations, \adarft can reduce RFT training time by up to 2$\times$, offering a practical path to more efficient RFT in structured reasoning tasks.

\section{Related Work}
\paragraph{Efficient Reinforcement Finetuning.} 
Most RFT pipelines build on Proximal Policy Optimization (PPO) \citep{schulman2017proximalpolicyoptimizationalgorithms}, with recent variants like RAFT \citep{dong2023raftrewardrankedfinetuning}, ReMax \citep{li2024remaxsimpleeffectiveefficient}, GRPO \citep{deepseekai2025deepseekr1incentivizingreasoningcapability}, and REINFORCE++ \citep{hu2025reinforcesimpleefficientapproach}, aiming to reduce computational overhead by simplifying RL components. While effective, these methods often trade off stability or sample efficiency. In parallel, data-centric strategies have emerged as promising alternatives for efficient finetuning. LIMO \citep{ye2025limoreasoning} and s1 \citep{muennighoff2025s1simpletesttimescaling} show that small, carefully selected supervised datasets can yield strong downstream performance, but their success hinges on manual curation, prompt engineering, and careful dataset construction, which may not generalize across tasks or models. LIMR \citep{li2025limrrlscaling} and \citet{wang2025reinforcementlearningreasoninglarge} proposes scoring training examples based on their estimated learning impact, enabling selective finetuning with fewer samples. Yet, computing these scores requires a full training run, and the scores must be recomputed for each new model, limiting practicality and scalability. Moreover, reducing the number of training samples does not inherently translate to improved efficiency. Models still require a comparable number of optimization steps and wall-clock time to converge. In contrast, \adarft introduces a lightweight, model-agnostic curriculum learning strategy that dynamically adjusts task difficulty based on reward feedback. This allows continuous adaptation to the model’s capabilities, improving convergence speed and final accuracy without modifying the RL algorithm or requiring manual data curation.

\paragraph{Curriculum Learning for RL.} 
Curriculum learning (CL) structures training by presenting tasks in an organized progression, typically from easy to hard, to enhance learning efficiency and generalization \citep{bengio2009curriculumlearning}. In RL, CL methods include task sorting by difficulty \citep{zaremba2015learningexecute, justesen2018illuminatinggeneralizationdeepreinforcement, wang2019pairedopenendedtrailblazerpoet}, teacher-student frameworks that adaptively select tasks based on learning progress \citep{matiisen2017teacherstudentcurriculumlearning, portelas2019teacheralgorithmscurriculumlearning}, and self-play approaches that induce automatic curricula through agent competition \citep{sukhbaatar2018intrinsicmotivationautomaticcurricula, zhao2025absolutezeroreinforcedselfplay}. Other strategies use intermediate-goal generation in sparse-reward settings \citep{florensa2018automaticgoalgenerationreinforcement}, unsupervised skill discovery \citep{jabri2019unsupervisedcurriculavisualmetareinforcement}, or knowledge transfer via progressive networks and imitation \citep{czarnecki2018mixmatchagentcurricula, rusu2022progressiveneuralnetworks}. While CL is well-studied in classical RL, its application to RFT of LLMs is still limited. Existing methods typically use staged training with hand-designed difficulty tiers \citep{wen2025lightr1curriculumsftdpo, deepscaler2025, song2025fastcurlcurriculumreinforcementlearning}, or online filtering schemes that repeatedly sample and discard data until rewards reach a target range \citep{bae2025online, yu2025dapoopensourcellmreinforcement}. These methods either lack adaptability or introduce significant computational overhead due to repeated rollouts. In contrast, \adarft is among the first truly adaptive curriculum learning approaches for RFT: it continuously adjusts task difficulty based on the model’s reward signal, enabling efficient, scalable training without fixed schedules or repeated rollouts.

\section{\adarft}
\label{sec:method}


We aim to improve the performance of a policy model $\pi_\theta$ for solving mathematical problems through adaptive curriculum learning. Fine-tuning on problems that are too easy or too hard leads to poor learning outcomes. Instead, the model should be trained on problems whose difficulty is close to the model's current capability. We frame this as an adaptive curriculum learning problem and propose \adarft, which adaptively adjusts the target difficulty to keep training problems within a suitable difficulty range. \adarft is compatible with a variety of RL algorithms (e.g, GRPO, PPO); in this work, we instantiate it with PPO and refer to this variant as \adarft(PPO).

Let $D$ be a dataset of mathematical problems, each annotated with a precomputed difficulty score $d_i$. The score can be either human-annotated or model-estimated.
The objective is to train a policy $\pi_\theta$ that improves its problem-solving ability by dynamically adjusting the training curriculum according to the model's current performance. Our proposed algorithm, \adarft, is shown in Algorithm~\ref{alg:curriculum}.

\begin{algorithm*}[t]
\small
\setlength{\baselineskip}{12pt}
\caption{\adarft -- Adaptive Curriculum Reinforcement Finetuning}
\label{alg:curriculum}
\begin{algorithmic}[1]
\STATE \textbf{Input:} Data source $D$ with difficulty scores $\{d_i\}$, policy model $\pi_\theta$, reward function $R(\cdot, \cdot)$, batch size $B$, initial target difficulty $T$, step size $\eta$, sensitivity $\alpha$, target reward $\beta$, difficulty bounds $d_{\min}, d_{\max}$
\STATE Select RL algorithm $\mathcal{A}$ (e.g., PPO, GRPO, REINFORCE++)
\WHILE{training is not finished}
    \STATE $\Delta_i \gets |d_i - T| \quad \forall i \in \{1, \ldots, |D|\}$ \hfill $\triangleright$ Compute differences between problem and target difficulty
    \STATE $X \gets \{ s_1, s_2, \ldots, s_B \}$ \hfill $\triangleright$ Sort and select top $B$ samples closest to target difficulty
    \STATE $G \gets \pi_\theta(X)$ \hfill $\triangleright$ Generate responses using policy model
    \STATE $R_{\text{avg}} \gets \frac{1}{|X|} \sum_{i=1}^{|X|} R(X_i, G_i)$ \hfill $\triangleright$ Compute average reward
    \STATE $\pi_\theta \gets \mathcal{A}(\pi_\theta, X, G, R)$ \hfill $\triangleright$ Update policy
    \STATE $T' \gets \text{clip}\!\left(T + \eta \cdot \tanh(\alpha \cdot (R_{\text{avg}} - \beta)), d_{\min}, d_{\max}\right)$ \hfill $\triangleright$ Update and clip target difficulty
    \STATE $T \gets T'$ \hfill $\triangleright$ Update sampler
\ENDWHILE
\end{algorithmic}
\end{algorithm*}

\subsection{Dynamic Curriculum Sampling}  
To construct an adaptive curriculum, we define a target difficulty $T$, which represents the current target difficulty level for training (more in Section~\ref{sec:difficulty}). \adarft dynamically adjusts $T$ based on the model’s reward signal to maintain an optimal difficulty level for learning. At each step, the algorithm computes the absolute difference between the target difficulty and the difficulty of each problem in the dataset (Alg.~\ref{alg:curriculum}, line 4):  
\( \Delta_i = |d_i - T| \) for all \( i \in [1, |D|] \). The batch of training problems is formed by selecting the $B$ problems with the smallest values of $\Delta_i$ (Alg.~\ref{alg:curriculum}, line 5), producing a batch:  
\( X = \{s_1, s_2, \ldots, s_B\} \). This deterministic nearest-difficulty selection is intentional: when the model cannot solve problems around the current target difficulty after many retry attempts, sampling substantially easier or harder problems provides a weaker curriculum signal for the target capability region. This ensures that the selected problems are closest to the model's current target difficulty, focusing the learning process on problems that are neither too easy nor too hard. A natural extension is to sample from a Gaussian or other smooth distribution centered at $T$, which may improve diversity while retaining the target-difficulty bias; we leave this exploration to future work.

\subsection{Policy Update}  
The selected batch $X$ is used to train the policy model $\pi_\theta$, which generates responses: \( G = \pi_\theta(X) \). A reward signal is computed based on the correctness of the model's output (Alg.~\ref{alg:curriculum}, line 7): $R_i = 1$ if the response is correct, and $R_i = 0$ if the response is incorrect.
The average reward over the batch is computed as (Alg.~\ref{alg:curriculum}, line 7):  
\( R_{avg} = \frac{1}{|X|} \sum_{i=1}^{|X|} R(X_i, G_i) \). The policy can then be updated using a reinforcement learning algorithm $\mathcal{A}$ such as PPO, GRPO, or REINFORCE++ (Alg.~\ref{alg:curriculum}, line 8):  
\( \pi_\theta \gets \mathcal{A}(\pi_\theta, X, G, R) \).

\subsection{Target Difficulty Update}  \label{sec:difficulty}
To adapt the curriculum dynamically, the target difficulty is updated based on the average reward. If the model performs well on the current difficulty level (high reward), the target difficulty increases, making the training problems harder. Conversely, if the model performs poorly, the target difficulty decreases. This dynamic update mechanism lies at the core of \adarft's curriculum adaptation strategy. The update rule (Alg.~\ref{alg:curriculum}, line 9) is defined as:  
\[
T' = \text{clip}(T + \eta \cdot \tanh(\alpha \cdot (R_{avg} - \beta)), d_{\min}, d_{\max}).
\]
The full derivation is provided in Appendix~\ref{sec:derive_target_update_rule}. This update rule is a smooth and stabilized variant of a standard linear mapping between reward space and difficulty space. Here, $\eta, \alpha, \beta$ are hyperparameters: $\eta$ is the step size for adjusting the target difficulty, $\alpha$ controls the sensitivity of the update, and $\beta$ is the target reward level, representing the desired success rate. The $\tanh$ function ensures smooth updates and prevents large jumps in difficulty by saturating for large deviations, while the ``clip'' function constrains the target difficulty within the valid range $[d_{\min}, d_{\max}]$. These bounds can be manually specified or automatically derived from the training set, for example, by taking the minimum and maximum of the difficulty scores $\{d_i\}$. Intuition and guidance for selecting these hyperparameters are discussed in Section \ref{sec:adarft-theory}, \ref{sec:experiment_training_setup}, and Appendix \ref{sec:derive_target_update_rule}.

\subsection{Motivation for Target Reward $\beta$}
\label{sec:adarft-theory}
A key component of \adarft is its adaptive curriculum mechanism, which steers training toward a target reward level $\beta$. Intuitively, we aim to train on examples that are neither trivially easy nor prohibitively hard. In this light, setting $\beta = 0.5$, corresponding to a success rate of roughly 50\%, naturally aligns with this goal. This section formalizes that intuition by analyzing the relationship between reward variance and learnability in RFT with binary rewards.

In entropy-regularized reinforcement learning, the optimal policy $\pi^*$ can be expressed relative to a reference policy $\pi_{\text{init}}$ as~\citep{korbak2022reinforcement, go2023aligning, rafailov2023direct}:
\begin{equation}
\pi^*(y \mid x) = \frac{1}{Z(x)}\pi_{\text{init}}(y \mid x)\exp\left(\frac{1}{\tau} r(x, y)\right)
\end{equation}
where $\tau$ is the inverse temperature parameter controlling entropy regularization, and $Z(x)$ is the partition function that normalizes the action probability. The corresponding optimal value function and the partition function is given by~\citep{schulman2017equivalence, richemond2024offline}:
\begin{equation}
V^*(x)
:= \tau \log \mathbb{E}_{y \sim \pi_{\text{init}}(\cdot \mid x)}
\left[ \exp\left(\frac{1}{\tau} r(x, y) \right) \right],
Z(x)
= \exp \left( \frac{1}{\tau} V^*(x) \right).
\end{equation}
We can then take the expectation of the log-ratio between the optimal policy and the initial policy with respect to $y \sim \pi_{\text{init}}(\cdot\mid x)$, leading to \citep{haarnoja2017reinforcement, schulman2017equivalence}:
\begin{equation}
\mathbb{E}_{y \sim \pi_{\text{init}}(\cdot \mid x)} 
\left[ \log \frac{\pi^*(y \mid x)}{\pi_{\text{init}}(y \mid x)} \right] = \frac{1}{\tau} \mathbb{E}_{\pi_{\text{init}}}\!\left[r(x, y)\right]
- \frac{1}{\tau} V^*(x).
\end{equation}

Since the left-hand side can be interpreted as the negative reverse KL divergence between $\pi_{\text{init}}$ and $\pi^*$ \citep{rafailov2024r}, \citet{bae2025online} show that when the reward $r(x, y)$ with $y \sim \pi_{\text{init}}(\cdot \mid x)$ is Bernoulli, the KL divergence is lower-bounded by the reward variance:
\begin{equation}
D_{\text{KL}}(\pi_{\text{init}} \| \pi^*) \geq \frac{p(x)(1 - p(x))}{2\tau^2}
\end{equation}
where $p(x)$ is the model's success rate on prompt $x$. This implies that the lower bound on the KL divergence is proportional to the reward variance, which is maximized when $p(x) = 0.5$. We use this result as an intuition for why prompts that the model succeeds on roughly half the time can provide informative policy-gradient updates; it does not by itself prove that PPO gradient norm or downstream learning progress is globally maximized at $p(x)=0.5$. In Section~\ref{sec:result} and Appendix~\ref{sec:ablation_beta}, we conduct an ablation study by varying the target reward $\beta$, demonstrating that setting $\beta = 0.5$ consistently leads to the best performance, empirically supporting the use of an intermediate-success curriculum.

\section{Experiments}
\label{sec:experiment}

\subsection{Difficulty Estimation}  
\label{sec:difficulty_estimation}

\begin{figure*}[t]
    \centering
    \begin{subfigure}[t]{0.32\textwidth}
        \centering
        \includegraphics[width=\textwidth]{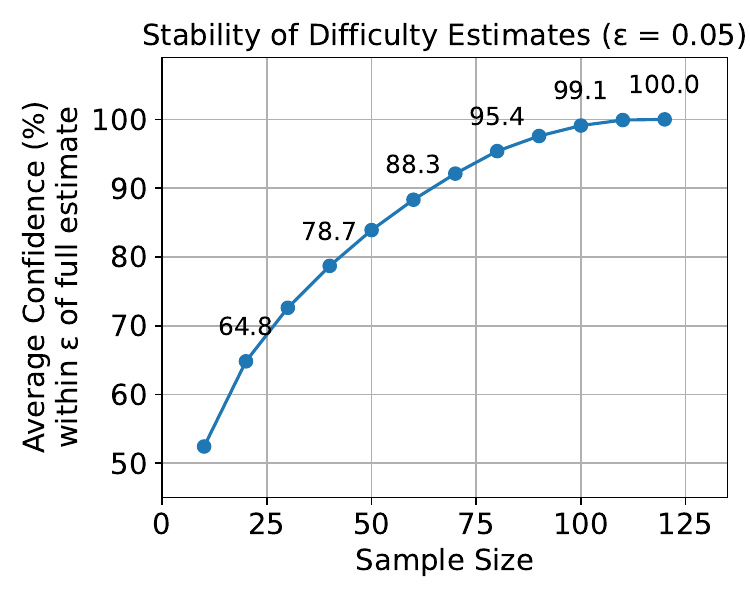}
        \caption{Average confidence that subsampled difficulty estimates fall within ±0.05 of the full-sample estimate.}
        \label{fig:difficulty_estimate}
    \end{subfigure}
    \hfill
    \begin{subfigure}[t]{0.66\textwidth}
        \centering
        \includegraphics[width=\textwidth]{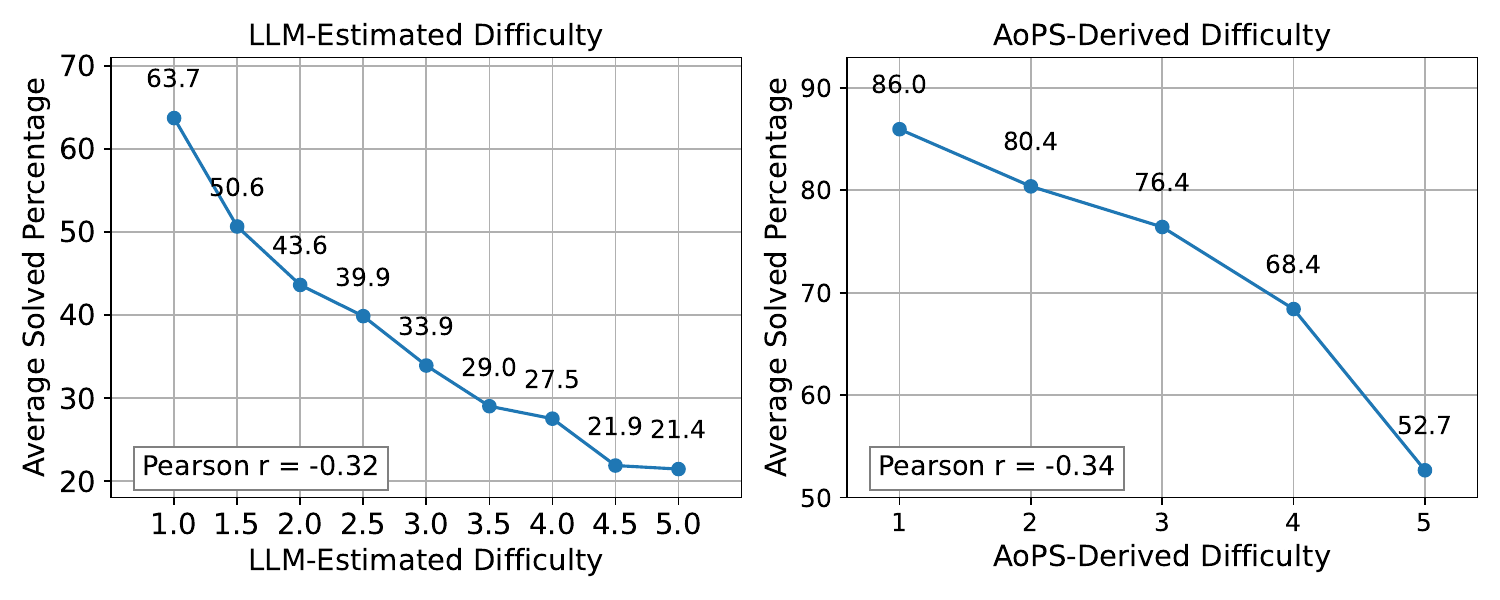}
        \caption{Correlation between average solved percentage and two types of difficulty labels: (left) LLM-estimated difficulty and (right) AoPS-derived difficulty levels.}
        \label{fig:correlation}
    \end{subfigure}
    \caption{Evaluation of difficulty estimation: (a) Stability of difficulty scores under subsampling of model rollouts; (b) Correlation between labeled difficulty levels and average solved percentage.}
    \label{fig:difficulty_eval}
\end{figure*}

Accurate estimation of problem difficulty is critical for \adarft. For difficulty estimation, we select the Qwen 2.5 MATH 7B model \citep{qwen2025qwen25technicalreport} because it demonstrates a balanced solving ability. A model that is too strong (e.g., OpenAI o1 \citep{openai2024openaio1card}, DeepSeek R1 \citep{deepseekai2025deepseekr1incentivizingreasoningcapability}) would solve most problems on the first attempt, leading to poor discrimination between easy and hard problems. Conversely, a model that is too weak (e.g., LLaMA 3.3 1B \citep{grattafiori2024llama3herdmodels}) would fail to solve most problems even after multiple attempts, limiting the signal required for curriculum adaptation. For each problem, the difficulty score is computed as: \( d_i = 100(1 - s_i/n) \), where \(s_i\) denotes the number of successful attempts on problem \(i\), and $n$ is the number of attempts per problem. In our setup, we use $n = 128$. 

To evaluate the stability of our difficulty estimation process, we simulate how confidence varies with different numbers of samples. For each problem, we treat the full set of 128 rollouts as the ground-truth difficulty estimate and compute how often sub-sampled estimates fall within a tolerance of $\epsilon = 0.05$. Specifically, we run 10 random sampling trials per sample size and average the confidence across all problems in the dataset. As shown in Figure \ref{fig:difficulty_estimate}, even with as few as 64 samples, the estimated difficulty remains within $\pm 0.05$ of the full estimate over 90\% of the time. With just 40 samples, the confidence remains around 80\%. These results indicate that accurate and robust difficulty estimation can be achieved with fewer rollouts, reducing the computational burden of large-scale curriculum construction. 

To further validate the reliability of our difficulty estimates, we examined their alignment with the difficulty levels provided in the MATH dataset. The MATH dataset comprises 12,500 competition-level mathematics problems sourced from contests such as the American Mathematics Competitions (AMC) and the American Invitational Mathematics Examination (AIME). Each problem is categorized into one of five difficulty levels, following the classification system used by the Art of Problem Solving (AoPS) community (details in Appendix \ref{sec:difficulty-estimation-prompt}). In this system, level 1 denotes the easiest problems, while level 5 represents the most difficult. As shown in Figure~\ref{fig:correlation}, there is a clear downward trend in the average solve rate as the labeled difficulty level increases, ranging from 86.0\% at level 1 to 52.7\% at level 5. Specifically, the AoPS-derived difficulty levels yield a Pearson correlation of $r = -0.34$ ($p < 0.05$) with model success rates. This negative correlation indicates that the model's empirical performance aligns well with the intended difficulty stratification, reinforcing the utility of both the labeled difficulty levels and our estimation approach in guiding curriculum learning. To further streamline the difficulty estimation process, we also prompted GPT-4o (\texttt{gpt-4o-0806}) \citep{openai2024gpt4ocard} to assign difficulty levels to the DeepScaleR dataset based on the AoPS rubric. Each problem was presented to GPT-4o with a request to rate its difficulty according to AoPS guidelines (the full prompt is shown in Appendix~\ref{sec:difficulty-estimation-prompt}). This approach provides a lightweight and scalable alternative to rollout-based estimation. As shown in Figure~\ref{fig:correlation}, GPT-4o’s difficulty ratings also correlate well with the  model success rates, with a Pearson correlation of $r = -0.32$ ($p < 0.05$), making it a practical proxy for curriculum scheduling when computational resources are constrained.

\subsection{Dataset}  
\label{sec:dataset}

\begin{figure*}[t]
    \centering
    \includegraphics[width=\textwidth]{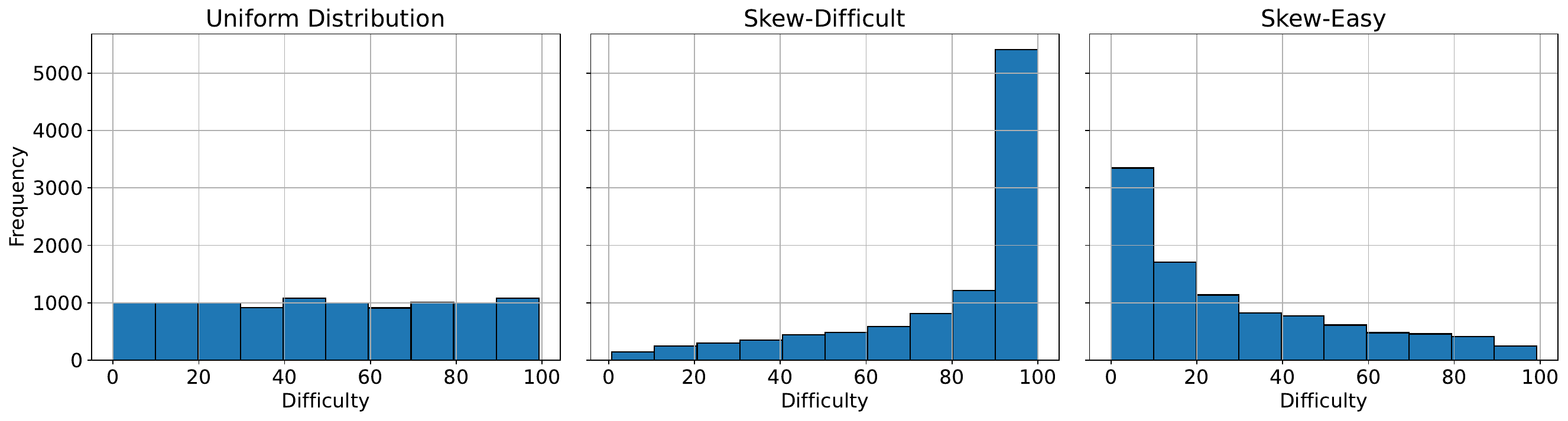}
    \caption{Difficulty distribution for different training sets: Uniform, Skew-Difficult, and Skew-Easy. Each training set contains 10,000 samples.}
    \label{fig:difficulty_distribution}
\end{figure*}

We use the DeepScaleR dataset \citep{deepscaler2025} as the training set. DeepScaleR compiles problems from multiple sources, including AIME from 1984 to 2023 and AMC prior to 2023. The dataset also includes problems from the Omni-MATH \citep{gao2024omnimathuniversalolympiadlevel} and Still datasets \citep{Slow_Thinking_with_LLMs_3_Preview}, which feature problems from various national and international math competitions. This results in a diverse and challenging training set, covering a wide range of mathematical domains and difficulty levels. 

In practice, we do not have control over the exact difficulty distribution of the data collected for training. This motivates our investigation into how different difficulty distributions influence \adarft. To this end, we construct three distinct distributions from the DeepScaleR dataset. The first is a \texttt{skew-difficult} distribution, where most problems are challenging. The second is a \texttt{skew-easy} distribution, where most problems are relatively easy. The third is a \texttt{uniform} distribution, where problems are evenly balanced across all difficulty levels, ensuring a consistent representation of easy, moderate, and hard problems. Each of these three distributions includes 10,000 samples. The data distribution for each setting is shown in Figure \ref{fig:difficulty_distribution}.

For evaluation, we use six benchmark datasets to assess the model’s performance across different levels of difficulty and mathematical reasoning. The first benchmark, MATH 500 \citep{lightman2023letsverifystepstep}, is a subset of the MATH dataset \citep{hendrycks2021measuring} containing 500 representative problems designed to test a model's general mathematical capability. GSM8K \citep{cobbe2021trainingverifierssolvemath} is a set of grade-school math problems. OlympiadBench \citep{he2024olympiadbenchchallengingbenchmarkpromoting} includes a collection of problems from Olympiad-level mathematics and physics competitions. Minerva Math \citep{lewkowycz2022solvingquantitativereasoningproblems} is a curated set of undergraduate-level math problems that assess complex mathematical reasoning and symbolic manipulation. AMC 23 and AIME 24 include problems from the 2023 American Mathematics Competitions and the 2024 American Invitational Mathematics Examination, respectively. Since AMC 23 contains only 40 problems and AIME 24 only 30, we report accuracy as the average over 8 sampled responses per problem to ensure stable estimates. Together, these datasets span elementary, high school, and advanced competition-level math, providing a comprehensive evaluation of the model’s reasoning abilities.

\begin{figure*}[t]
    \centering
    \includegraphics[width=\textwidth]{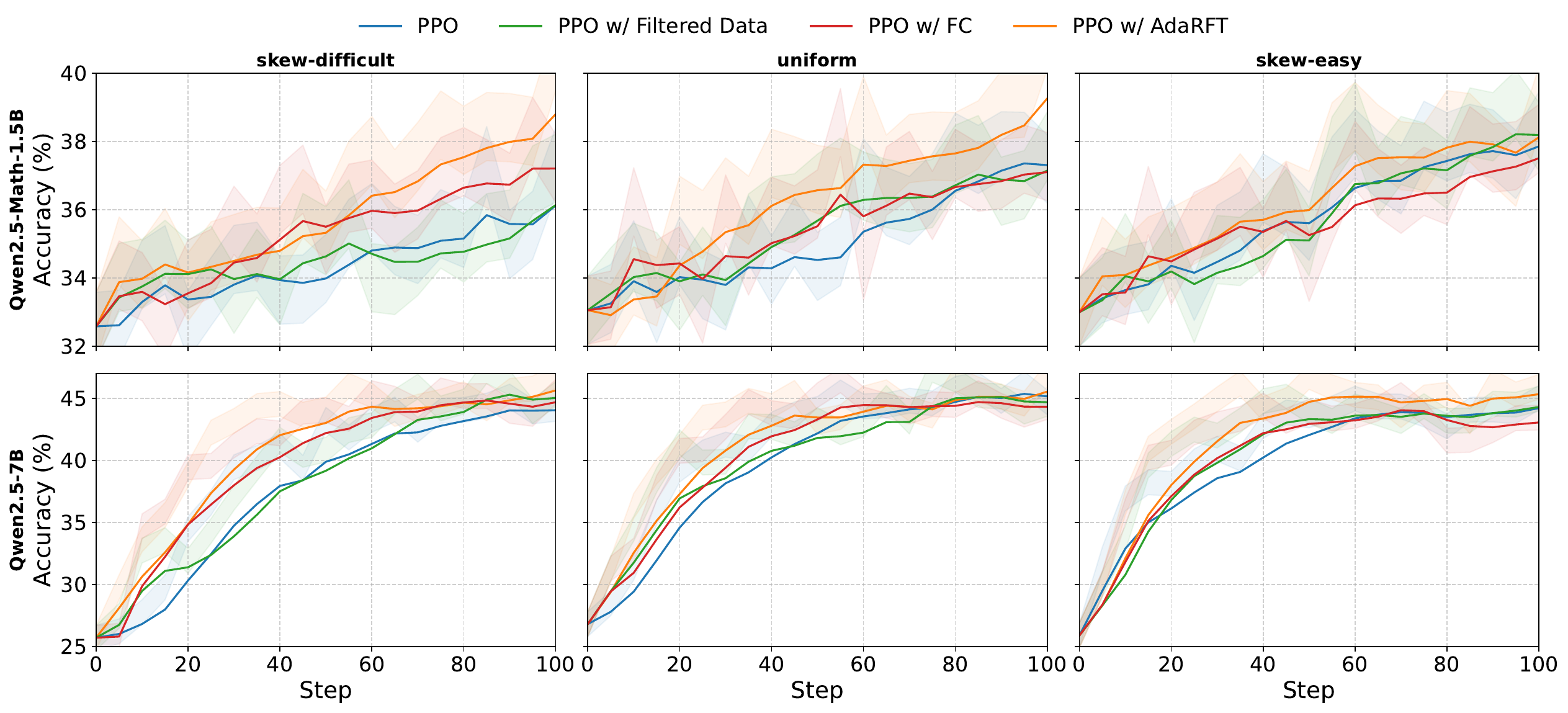}
    \caption{Performance comparison of PPO, PPO with filtered data, \textcolor{black}{PPO with fixed curriculum (PPO w/ FC)}, and \adarft(PPO) across different setups (uniform, skew-easy, skew-difficult). Accuracy is the average of MATH 500, GSM8K, AIME 24, AMC 23, OlympiadBench, and Minerva Math. Compared with baselines, \adarft improves both the accuracy and training efficiency. For clarity, curves are exponentially smoothed.}
    \label{fig:result}
\end{figure*}

\subsection{Training Setup}
\label{sec:experiment_training_setup}
We trained two models on the three difficulty-based distributions of the DeepScaleR dataset described in Section \ref{sec:dataset}: Qwen 2.5 7B and Qwen 2.5 MATH 1.5B. This setup allows us to evaluate the effectiveness of \adarft on models with different initial performance levels when exposed to skew-difficult, skew-easy, and uniform problem distributions. All models were trained using four different approaches: (1) the standard PPO algorithm, (2) \adarft(PPO), our method that integrates adaptive curriculum learning with PPO (see Section~\ref{sec:method}), (3) PPO with filtered data, a baseline that trains PPO on data filtered by pass@$k$ accuracy, and (4) PPO with a fixed curriculum schedule.

For the data filtering baseline (3), following prior work \citep{bae2025online, hu2025openreasonerzeroopensourceapproach, zr1}, we first run a pass@40 analysis for each combination of model and data distribution. We then discard examples that are either too easy or too hard, removing all problems with solved rates $\leq 10\%$ or $\geq 90\%$. This restricts training to problems of intermediate difficulty. However, this procedure removes a large fraction of the data, including many potentially informative examples. In addition, because difficulty is defined using pass@$k$ metrics, the filtering must be recomputed whenever the model or the data distribution changes.

\textcolor{black}{For the fixed curriculum baseline (4), we follow the approach of prior work \citep{parashar2025curriculumreinforcementlearningeasy, kimiteam2025kimik15scalingreinforcement}. In this setting, the difficulty of sampled problems follows a predetermined schedule that increases linearly over training steps. Suppose training runs for $m$ total steps. We then define a target difficulty $T(s)$ at step $s$ by
\(
T(s) = T_{\min} + \frac{T_{\max} - T_{\min}}{m} \cdot s ,
\)
and at each step we sample problems whose estimated difficulty matches this target. Unlike \adarft, this schedule increases difficulty at a fixed rate regardless of how quickly or slowly the model learns.}


The training batch size was set to $B = 1024$, with the target reward $\beta$ set to 0.5 to promote learning at a balanced success rate. The sensitivity parameter $\alpha$ and step size $\eta$ were tuned using a validation set to ensure stable curriculum updates. We set $\alpha = 2$, $\eta = 50$, and the initial target difficulty $T = 0$. The step size $\eta$ acts as a scaling factor between the reward signal and the difficulty metric. Since the difficulty metric ranges from 0 to 100 and the reward ranges from 0 to 1, a target reward $\beta = 0.5$ implies that the maximum reasonable adjustment to the difficulty metric should be around 50. Therefore, we set $\eta = 50$ to scale the reward signal appropriately to the difficulty range. The sensitivity parameter $\alpha = 2$ controls the slope of the $\tanh$ function. Setting $\alpha$ to 2 makes the $\tanh$ function behave approximately linearly when the difference between the average reward and the target reward is small. The $\tanh$ function smooths out the adjustments, allowing for more controlled changes when the difference is large while maintaining sensitivity when the difference is small. Both models were trained on 8 A100 GPUs for approximately 100 steps. The implementation details can be found in Appendix \ref{sec:implementation_details}.

\section{Results and Analysis}
\label{sec:result}

We evaluate the performance of standard PPO and \adarft(PPO) across multiple training setups and two model sizes: Qwen 2.5 MATH 1.5B and Qwen 2.5 7B. Figure~\ref{fig:result} presents the learning curves averaged across six benchmarks, while Table \ref{tab:result_step60} and \ref{tab:result_step100} provide a detailed breakdown of accuracy and training efficiency. On average, models trained with \adarft(PPO) reach comparable accuracy in fewer RFT training steps and also obtain higher final accuracy in our single-seed runs. This improvement is particularly notable in non-uniform data distributions, where curriculum adaptation is most beneficial.

\begin{table*}[t]
\centering
\resizebox{\textwidth}{!}{%
\small
\begin{tabular}{lllcccccc}
\toprule
Model & Setup & Method & Final Acc (\%) $\uparrow$ & Avg Step Time (s) $\downarrow$ & Extra Steps (\%) $\downarrow$ & Extra Steps $\downarrow$ & Extra Time (s) $\downarrow$ \\
\midrule
\multirow{12}{*}{\begin{tabular}{c}Qwen2.5\\Math\\1.5B\end{tabular}}
 & \multirow{4}{*}{skew-difficult}
   & \adarft & \textbf{40.48} & \textbf{122.24} & \textbf{0.0\%} & \textbf{+0} & \textbf{0.00} \\
 &   & PPO & 37.41 & 132.95 & 71.7\% & +43 & 5716.85 \\
 &   & PPO (w/ Filter) & 37.22 & 128.20 & 81.7\% & +49 & 6281.80 \\
 &   & \textcolor{black}{PPO (w/ FC)} & 37.22 & \textcolor{black}{130.91} & \textcolor{black}{26.7\%} & \textcolor{black}{+16} & \textcolor{black}{2094.56} \\
\cmidrule(lr){2-8}
 & \multirow{4}{*}{uniform}
   & \adarft & \textbf{41.11} & \textbf{121.31} & \textbf{0.0\%} & \textbf{+0} & \textbf{0.00} \\
 &   & PPO & 37.20 & 126.82 & 56.7\% & +34 & 4311.88 \\
 &   & PPO (w/ Filter) & 37.87 & 126.35 & 86.7\% & +52 & 6570.20 \\
 &   & \textcolor{black}{PPO (w/ FC)} & 37.27 & \textcolor{black}{126.40} & \textcolor{black}{80.0\%} & \textcolor{black}{+48} & \textcolor{black}{6067.20} \\
\cmidrule(lr){2-8}
 & \multirow{4}{*}{skew-easy}
   & \adarft & \textbf{39.18} & 120.52 & \textbf{0.0\%} & \textbf{+0} & \textbf{0.00} \\
 &   & PPO & 38.46 & 121.15 & 26.7\% & +16 & 1938.40 \\
 &   & PPO (w/ Filter) & 38.15 & \textbf{115.12} & 35.0\% & +21 & 2417.52 \\
 &   & \textcolor{black}{PPO (w/ FC)} & 38.09 & \textcolor{black}{121.99} & \textcolor{black}{58.3\%} & \textcolor{black}{+35} & \textcolor{black}{4269.65} \\
\midrule
\multirow{12}{*}{\begin{tabular}{c}Qwen2.5\\7B\end{tabular}}
 & \multirow{4}{*}{skew-difficult}
   & \adarft & \textbf{46.83} & \textbf{239.92} & \textbf{0.0\%} & \textbf{+0} & \textbf{0.00} \\
 &   & PPO & 44.17 & 246.21 & 60.0\% & +24 & 5909.04 \\
 &   & PPO (w/ Filter) & 45.35 & 254.22 & 62.5\% & +25 & 6355.50 \\
 &   & \textcolor{black}{PPO (w/ FC)} & 45.54 & \textcolor{black}{243.12} & \textcolor{black}{22.5\%} & \textcolor{black}{+9} & \textcolor{black}{2188.08} \\
\cmidrule(lr){2-8}
 & \multirow{4}{*}{uniform}
   & \adarft & \textbf{46.92} & \textbf{234.16} & \textbf{0.0\%} & \textbf{+0} & \textbf{0.00} \\
 &   & PPO & 44.70 & 243.82 & 32.5\% & +13 & 3169.66 \\
 &   & PPO (w/ Filter) & 44.57 & 263.11 & 57.5\% & +23 & 6051.53 \\
 &   & \textcolor{black}{PPO (w/ FC)} & 44.32 & \textcolor{black}{240.62} & \textcolor{black}{17.5\%} & \textcolor{black}{+7} & \textcolor{black}{1684.34} \\
\cmidrule(lr){2-8}
 & \multirow{4}{*}{skew-easy}
   & \adarft & \textbf{45.94} & 247.44 & \textbf{0.0\%} & \textbf{+0} & \textbf{0.00} \\
 &   & PPO & 45.07 & 235.27 & 50.0\% & +20 & 4705.40 \\
 &   & PPO (w/ Filter) & 44.98 & \textbf{233.13} & 42.5\% & +17 & 3963.21 \\
 &   & \textcolor{black}{PPO (w/ FC)} & 43.46 & \textcolor{black}{240.66} & \textcolor{black}{57.5\%} & \textcolor{black}{+23} & \textcolor{black}{5535.18} \\
\bottomrule
\end{tabular}
}
\caption{Average training time per step (in seconds), final average accuracy (\%) at step 100 averaged over MATH 500, GSM8K, AIME 24, AMC 23, OlympiadBench, and Minerva Math, and the additional training steps required for each baseline to match \adarft’s accuracy at step 60 for Qwen 2.5 Math 1.5B or step 40 for Qwen 2.5 7B, across different setups and methods.}
\label{tab:result_step60}
\end{table*}

\subsection{Training Efficiency}

As shown in Figure~\ref{fig:result} and Table~\ref{tab:result_step60}, models trained with \adarft consistently require fewer training steps to match the performance of those trained with standard PPO, PPO on filtered data, and PPO with a fixed curriculum schedule. Specifically, we report how many additional steps are needed for PPO variants to match the performance of \adarft at step 60 for Qwen 2.5 Math 1.5B, and step 40 for Qwen 2.5 7B. Because models are evaluated only every 5 training steps, we apply exponential smoothing with a smoothing parameter of $0.3$ to the accuracy curves to reduce variance. The shaded areas in Figure~\ref{fig:result} represent the raw, unsmoothed accuracy $\pm$1\%. For Qwen 2.5 Math 1.5B, standard PPO requires 43 extra steps (+71.7\%) in the skew-difficult setting and 34 steps (+56.7\%) in the uniform setting to match \adarft's performance. PPO with filtered training data requires even more: +49 steps (81.7\%) and +52 steps (86.7\%) in the respective settings. In the skew-easy scenario, PPO requires +16 steps (26.7\%), while PPO with filtered data needs +21 steps (35.0\%) to catch up to \adarft. The efficiency gains remain substantial with the larger Qwen 2.5 7B model. In the skew-difficult setting, PPO and PPO with filtered data require +24 steps (60.0\%) and +25 steps (62.5\%), respectively. \textcolor{black}{PPO with a fixed curriculum schedule also follows this trend, suggesting that while fixed curricula can modestly improve training efficiency, their inability to adapt the the model’s evolving learning dynamics limits their convergence speed relative to \adarft.}

In addition to improved sample efficiency, \adarft also achieves faster average training time per step across nearly all settings, as reported in Table~\ref{tab:result_step60}. 
This is largely due to the fact that easier problems require fewer tokens to solve. For example, an arithmetic reasoning question from GSM8K might require only around 200 tokens to reach a correct answer, whereas a competition-level math problem from AIME could require around 2000 tokens, a 10$\times$ difference in rollout length. 
The total token length affects multiple components of the training step, including the rollout itself and the subsequent PPO update. While PPO update time does not scale linearly with sequence length due to batching and attention computation patterns, longer sequences still incur higher compute costs. 
As a result, curriculum learning’s tendency to prioritize easier problems early in training leads to shorter sequences on average, reducing per-step compute and improving overall training throughput.

\subsection{Model Performance}

While the primary goal of \adarft is to improve training efficiency, it also consistently maintains and often improves final model performance in our runs. As shown in Tables \ref{tab:result_step60}, \ref{tab:result_step100}, and Figure \ref{fig:result}, \adarft achieves higher final accuracy at the end of training (step 100) across all configurations. Reported results are averaged over six benchmarks: GSM8K, MATH 500, OlympiadBench, Minerva Math, AMC 23, and AIME 24. On skew-difficult data, Qwen 2.5 Math 1.5B improves from 37.41\% with PPO to 40.48\% with \adarft, a gain of over 3 percentage points. Similar gains are observed in the uniform setting (41.11\% vs.\ 37.20\%). Even on skew-easy data, where the baseline already performs well, \adarft still yields modest improvements (39.18\% vs.\ 38.46\%). For the larger Qwen 2.5 7B model, improvements are smaller but consistent: from 44.17\% to 46.83\% on skew-difficult data, from 44.70\% to 46.92\% in the uniform setting, and from 45.07\% to 45.94\% on skew-easy data. Since these final accuracy results are based on one seed per configuration, they should be interpreted as empirical trends rather than statistically significant accuracy improvements. They support the conclusion that \adarft improves training efficiency without sacrificing final accuracy. A more detailed per-benchmark breakdown is provided in Appendix \ref{sec:model_performance_details}.

\subsection{Ablation on Target Reward $\beta$}
\label{sec:ablation_beta}

\begin{figure*}[t]
  \centering
  \includegraphics[width=\textwidth]{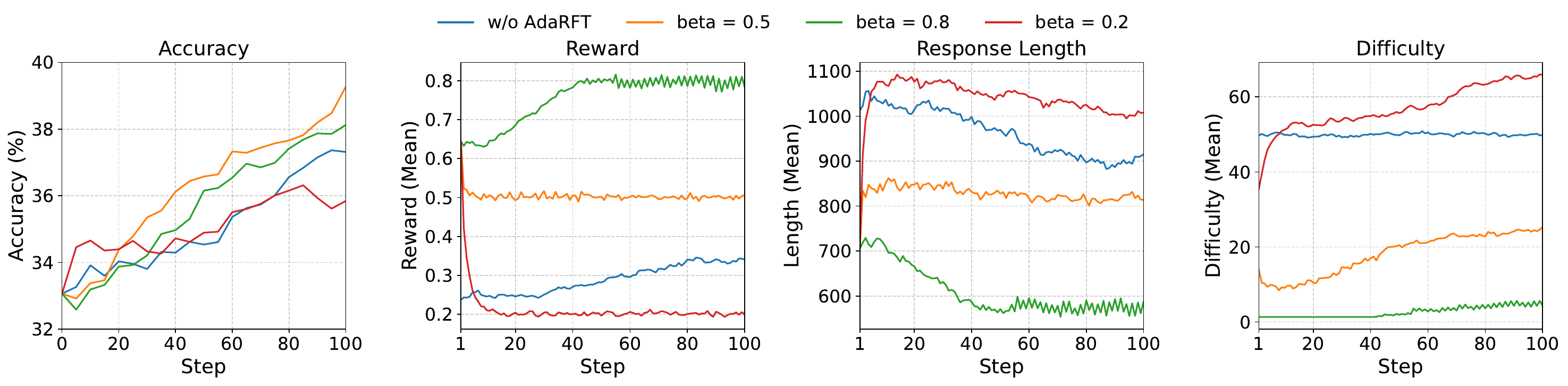}
  \caption{Ablation on $\beta$ in \adarft: we compare model accuracy, average reward, response length, and mean difficulty under $\beta=0.2$, $\beta=0.5$, and $\beta=0.8$, along with standard PPO (w/o \adarft).}
  \label{fig:beta_ablation}
\end{figure*}

To better understand the role of the target reward $\beta$ in \adarft, we perform an ablation study varying $\beta$ in the target difficulty update rule. Recall that $\beta$ controls the target average reward the model is expected to achieve and implicitly steers the curriculum: lower values prioritize easier problems, while higher values shift the curriculum toward more challenging samples. We train a Qwen 2.5 Math 1.5B model on the uniform data distribution with \adarft (PPO) using three different values of $\beta$: 0.2, 0.5, and 0.8. For comparison, we also include standard PPO without \adarft (denoted as ``w/o \adarft'') as a baseline.

As shown in Figure~\ref{fig:beta_ablation}, the model trained with $\beta = 0.5$ achieves the highest accuracy throughout training. This supports our theoretical motivation in Section \ref{sec:adarft-theory}: maximizing reward variance, which occurs when success rate $\approx 0.5$, provides the strongest learning signal. Models with $\beta = 0.2$ and $\beta = 0.8$ underperform likely due to curriculum misalignment: $\beta = 0.8$ overly focuses on easy problems, while $\beta = 0.2$ overemphasizes difficult ones, both of which limit the model’s capacity to generalize. The reward and difficulty curves align with the accuracy outcomes discussed above. The $\beta = 0.5$ configuration maintains a stable reward near 0.5, reflecting balanced difficulty exposure. In contrast, $\beta = 0.8$ results in overly high reward (i.e., easy samples), while $\beta = 0.2$ maintains a reward around 0.2 for most of training, indicating the model is repeatedly presented with overly difficult problems. As expected, response length is the shortest for $\beta = 0.8$ and longest for $\beta = 0.2$, consistent with the idea that longer responses correlate with problem complexity.

\subsection{Data Difficulty on Model Performance}
\label{sec:adarft_data_difficulty}
\begin{figure*}[ht]
    \centering
    \includegraphics[width=\textwidth]{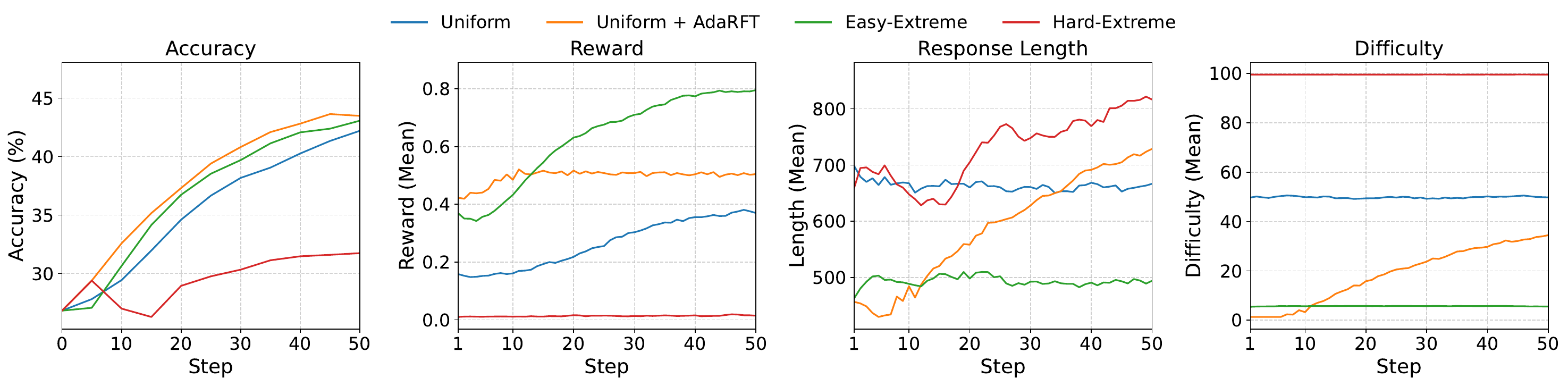}
    \caption{Performance comparison of Qwen 2.5 7B trained on different data distributions using PPO (Uniform, Easy-Extreme, Hard-Extreme) and \adarft instantiated with PPO (Uniform + \adarft). For clarity, curves are exponentially smoothed ($\alpha = 0.3$) to reduce noise.}
    \label{fig:extreme_result}
\end{figure*}

To better understand the effect of data difficulty on model performance, we introduce two additional data distributions: easy-extreme and hard-extreme. Unlike the skew-difficult and skew-easy distributions, which still include a mix of difficulty levels, the easy-extreme and hard-extreme sets consist exclusively of the most polarized examples. Specifically, easy-extreme contains only the easiest samples with difficulty levels no greater than 15, while hard-extreme includes only the hardest samples with difficulty levels of at least 97. Each of these extreme distributions consists of approximately 8,000 samples, providing a focused and controlled evaluation of model behavior under minimal or maximal difficulty conditions. We trained a Qwen 2.5 7B model on each of the two extreme distributions using PPO, and compared their performance to models trained on the uniform distribution with PPO (Uniform) and with \adarft instantiated with PPO (Uniform + \adarft), as described in Section \ref{sec:result}. The results are presented in Figure \ref{fig:extreme_result}. The key takeaway is that training on only overly easy or hard problems fails to provide useful learning signals, reinforcing the need for \adarft to adaptively steer models toward challenges matched to their current ability.

\paragraph{Accuracy.} The leftmost panel of Figure \ref{fig:extreme_result} shows that uniform + \adarft achieves the highest overall accuracy throughout training, outperforming both uniform and the two extreme settings. This highlights the effectiveness of \adarft in guiding the model through an optimal difficulty progression. In contrast, hard-extreme struggles, with a flat and lower trajectory, indicating that exposing the model only to very difficult problems limits learning progress. This suggests that without a gradual exposure strategy, models trained on only the hardest problems are unable to bootstrap their capabilities effectively.

\paragraph{Reward.} The reward trends provide important clues about learning dynamics. The easy-extreme setup achieves the fastest reward improvement during early training, surpassing both uniform and hard-extreme. In particular, easy-extreme consistently operates in a reward range between 0.4 and 0.6 during early training, which corresponds to a success rate that is both challenging and attainable. In contrast, the reward of the uniform and hard-extreme setup lingers below 0.2 in early training, leading to slower learning. This suggests that training on problems with intermediate difficulty—those that are neither trivially easy nor prohibitively hard—provides the most effective learning signal.  Notably, \adarft is explicitly designed to exploit this insight: by setting the target reward $\beta = 0.5$, we encourage the model to train on problems that match this ``productive struggle'' zone. As shown by the uniform + \adarft curve, the algorithm successfully maintains an average reward near 0.5 throughout training, allowing the model to learn at an optimal pace. Notably, while the uniform setup eventually reaches a reward of nearly 0.5 by step 50, it does not result in faster learning. This is likely because the model is already fairly well trained by that stage, so the additional reward signal contributes less to further improvement. In contrast, the hard-extreme model receives almost no reward signal for most of the training, while the uniform setup shows slower and more gradual reward accumulation.

\paragraph{Response Length.} The response length panel reveals how the complexity of generated solutions evolves during training. The hard-extreme model consistently produces the longest responses, with length increasing steadily, reflecting the higher complexity and reasoning depth required by the hardest problems. In contrast, the easy-extreme setup maintains short and stable responses, consistent with its simpler problem set. The uniform and uniform + \adarft setups fall between these two extremes. Notably, uniform + \adarft shows a gradual increase in response length over time. This trend aligns with the behavior of the curriculum learning algorithm: as the model improves, it is exposed to increasingly difficult problems, which naturally demand more elaborate reasoning and longer solutions. This dynamic suggests that response length can serve as a useful proxy for problem difficulty and reasoning complexity during training.

\paragraph{Difficulty.} Finally, the difficulty panel illustrates how problem difficulty evolves under each setup. The easy-extreme and hard-extreme curves remain flat, confirming that these datasets contain only problems from the tail ends of the difficulty spectrum (i.e., $\leq 15$ and $\geq 97$, respectively). The uniform curve is centered around 50, as expected, while uniform + \adarft shows a steady increase in difficulty over time. This adaptive progression confirms that curriculum learning effectively steers the model from easier to harder problems, aligning difficulty with the model’s evolving capabilities.

\subsection{\adarft with Diverse RL Algorithms}
\label{sec:adarft_rl}

\begin{wrapfigure}{r}{0.4\textwidth}
    \centering
    \vspace{-1em}
    \includegraphics[width=\linewidth]{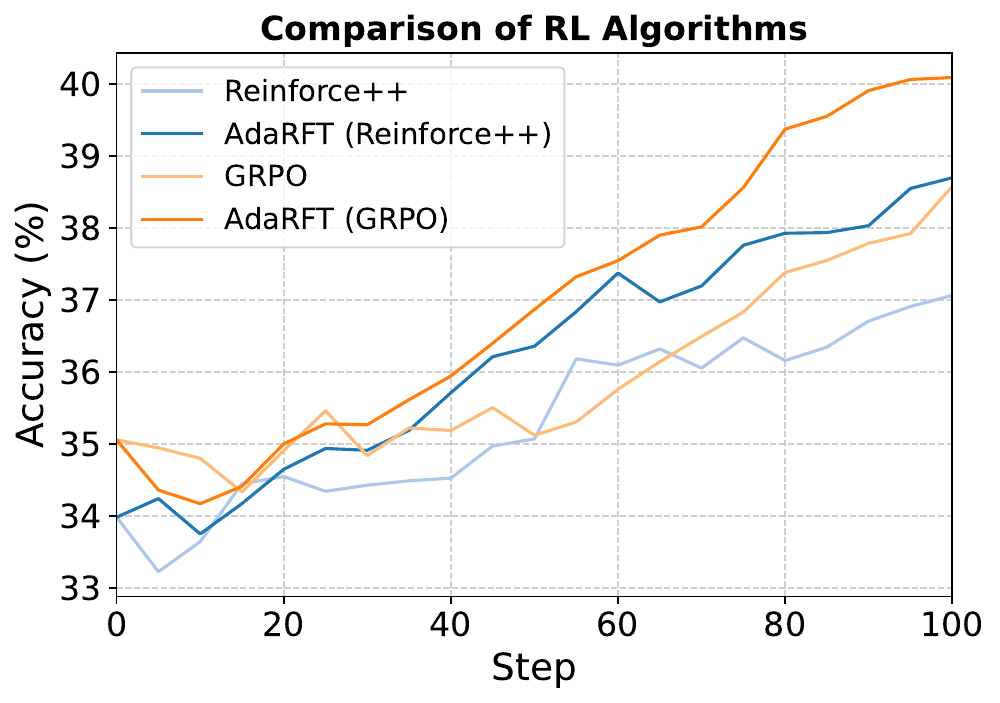}
    \caption{Comparison between models trained with and without AdaRFT using REINFORCE++ and GRPO.}
    \label{fig:optim_strategies_comparison}
    \vspace{-1em}
\end{wrapfigure}

To evaluate the generality of \adarft beyond PPO, we trained the Qwen 2.5 Math 1.5B model on a skew-difficult data distribution using two alternative reinforcement learning algorithms: REINFORCE++ and GRPO (implementation details in Appendix \ref{sec:implementation_details}). As shown in Figure~\ref{fig:optim_strategies_comparison}, \adarft improves convergence speed and final accuracy across these variants in our single-seed runs. Across both cases, the adaptive curriculum acts orthogonally to the underlying optimization method. These results reinforce the plug-and-play nature of \adarft: it consistently enhances sample efficiency across algorithmic choices, making it broadly applicable in diverse reinforcement finetuning pipelines. Notably, this generalization holds without any additional tuning or algorithm-specific modifications, underscoring the practical utility of curriculum-aware training in both lightweight and computation-heavy RFT settings.

\subsection{Training on LLM-Estimated Difficulty}
\label{sec:adarft_llm}

\begin{wrapfigure}{r}{0.4\textwidth}
  \centering
  \vspace{1em}
  \includegraphics[width=0.38\textwidth]{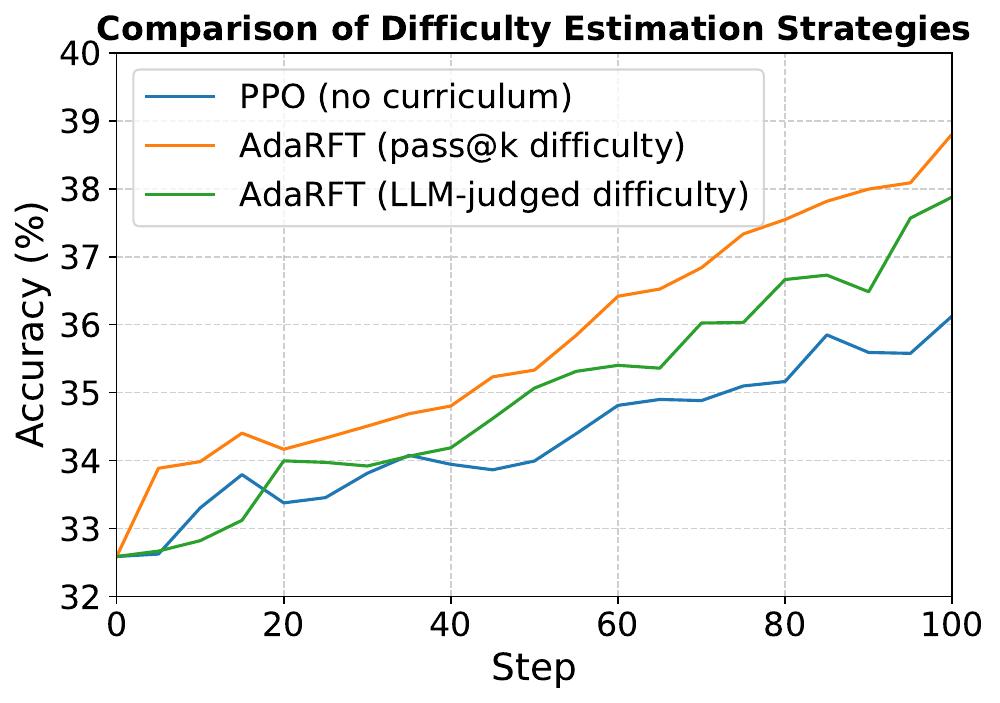}
  \vspace{-10pt}
  \caption{Comparison of different difficulty estimation strategies.}
  \label{fig:llm_difficulty_training}
  \vspace{-1em}
\end{wrapfigure}

In addition to rollout-based difficulty estimation, we explore an alternative strategy that uses LLM-judged difficulty levels to guide curriculum construction. As described in Section \ref{sec:difficulty_estimation}, we prompt GPT-4o (\texttt{gpt-4o-0806}) to assign difficulty levels to math problems in the DeepScaleR dataset according to the AoPS rubric. This approach offers a lightweight and scalable alternative to computing pass@$k$ success rates from model rollouts, making it especially attractive in low-resource scenarios.

To assess the effectiveness of this strategy, we train a Qwen 2.5 Math 1.5B model on the skew-difficult distribution using \adarft (PPO) with two curriculum schedules: one based on rollout-derived pass@$k$ difficulty, and the other guided by GPT-4o's difficulty ratings. Since the LLM-judged difficulty is on a scale of 1 to 5 (rather than 0 to 100), we set the step size hyperparameter \(\eta = 2.5\) to align the difficulty adjustment magnitude with the reward signal. All other hyperparameters are kept unchanged. As shown in Figure~\ref{fig:llm_difficulty_training}, both curriculum strategies outperform standard PPO without curriculum learning. While rollout-based difficulty estimation yields the strongest gains, the LLM-judged curriculum still provides a noticeable improvement over the baseline.

These results demonstrate that \adarft remains effective even when the difficulty signal is derived from heuristic or approximate sources like LLM judgments. Although less precise than empirical pass@$k$ metrics, the LLM-based difficulty still provides enough structure to enable meaningful curriculum adaptation. This makes it a practical fallback when rollout computation is too costly, and suggests that future work could explore hybrid approaches that combine lightweight heuristics with periodic empirical calibration.

\section{Discussion}
\label{sec:discussion}

\paragraph{When Does Curriculum Learning Help Most?}
Our findings show that curriculum learning provides the greatest benefits under two key conditions: (1) imbalanced training distributions, and (2) limited model capacity. In skewed settings, standard PPO often struggles early due to sparse reward signals. \adarft mitigates this issue by prioritizing easier problems, allowing the model to bootstrap learning before progressing to harder ones. The gains are smaller when the model is strong or the data is well-balanced, since the training distribution is already sufficiently informative. Manual data curation and fixed curriculum schedules can partially address these issues \citep{yu2025dapoopensourcellmreinforcement, shen2025exploringdatascalingtrends, chen2025empiricalstudyelicitingimproving, zeng2025simplerlzooinvestigatingtamingzero}, but they require substantial human effort and domain expertise and often need to be re-tuned for each model or task. Fixed curricula also cannot adapt to differences in learning speed. We provide a detailed analysis of fixed curricula in Appendix~\ref{sec:training_dynamics_fc}. In contrast, \adarft dynamically adjusts difficulty based on the model’s reward signal without manual intervention. This makes it broadly applicable across model scales and data distributions, improving scalability and robustness, especially in fixed data settings where it can adapt to the model without modifying the dataset.

\paragraph{Use of Offline Difficulty Estimation.}
A core design choice of \adarft is to rely on offline difficulty estimates that remain fixed throughout training, rather than recomputing difficulty online as the model improves. \adarft adopts a solver-independent notion of difficulty, in which problems such as IMO-level questions are considered inherently more difficult than grade-school math problems and are assigned a single difficulty score based on intrinsic complexity, in contrast to online curriculum methods that use solver-dependent definitions that evolve with the model’s success rate. This design aligns naturally with many mathematics competitions~\citep{hendrycks2021measuring} and puzzle datasets~\citep{lin2025zebralogic}, which provide fixed difficulty annotations, and with modern RL fine-tuning pipelines where pass@$n$ rollouts are commonly computed during data preparation regardless of curriculum learning. Recent systems such as Open Reasoner Zero~\citep{hu2025openreasonerzeroopensourceapproach}, Kimi k1.5~\citep{kimiteam2025kimik15scalingreinforcement}, and ZR1 1.5B~\citep{zr1} evaluate pass@$n$ on the full dataset prior to training to filter overly easy or difficult examples, making difficulty information readily available in practice. Rather than updating difficulty estimates themselves, \adarft adapts the curriculum by dynamically updating the target difficulty $T$ based on the model’s success rate on the currently sampled data, achieving the same functional outcome as online difficulty estimation while remaining stable and scalable for large-scale RL fine-tuning.

We report curriculum construction and policy optimization as separate sources of compute. The efficiency results in Table~\ref{tab:result_step60} measure the RFT training phase after the curriculum scores have been assigned; they do not include the upfront difficulty-estimation pass. In our main rollout-based setup, assigning difficulty to a 10,000-example training pool with pass@128 requires 1.28M scorer rollouts. This cost should be included explicitly for one-off training runs, and our 2$\times$ speedup claim should therefore be interpreted as a reduction in the subsequent RFT training phase. The cost can be reduced by using fewer rollouts, since Figure~\ref{fig:difficulty_estimate} shows that pass@40 and pass@64 preserve much of the pass@128 ordering, or by using LLM-judged difficulty as in Section~\ref{sec:adarft_llm}. It can also be amortized when the same scored dataset is reused across model sizes, algorithms, hyperparameter sweeps, or future training runs.

\section{Conclusion}
We propose \adarft, an adaptive curriculum learning strategy for reinforcement finetuning (RFT) that dynamically matches problem difficulty to a model’s evolving skill level. By adjusting a target difficulty based on reward feedback, \adarft improves sample and compute efficiency during RFT training with problem-level difficulty annotations, without modifying the reward function or underlying RL algorithm. Experiments across multiple data regimes and model sizes show consistent gains in convergence speed, especially in imbalanced training distributions. This lightweight, scalable approach highlights the value of curriculum-aware training for efficient and robust alignment in structured reasoning tasks.




\clearpage
\bibliography{main}

@misc{kazemnejad2024vineppounlockingrlpotential,
      title={VinePPO: Unlocking RL Potential For LLM Reasoning Through Refined Credit Assignment}, 
      author={Amirhossein Kazemnejad and Milad Aghajohari and Eva Portelance and Alessandro Sordoni and Siva Reddy and Aaron Courville and Nicolas Le Roux},
      year={2024},
      eprint={2410.01679},
      archivePrefix={arXiv},
      primaryClass={cs.LG},
      url={https://arxiv.org/abs/2410.01679}, 
}

@misc{dong2023raftrewardrankedfinetuning,
      title={RAFT: Reward rAnked FineTuning for Generative Foundation Model Alignment}, 
      author={Hanze Dong and Wei Xiong and Deepanshu Goyal and Yihan Zhang and Winnie Chow and Rui Pan and Shizhe Diao and Jipeng Zhang and Kashun Shum and Tong Zhang},
      year={2023},
      eprint={2304.06767},
      archivePrefix={arXiv},
      primaryClass={cs.LG},
      url={https://arxiv.org/abs/2304.06767}, 
}

@misc{jabri2019unsupervisedcurriculavisualmetareinforcement,
      title={Unsupervised Curricula for Visual Meta-Reinforcement Learning}, 
      author={Allan Jabri and Kyle Hsu and Ben Eysenbach and Abhishek Gupta and Sergey Levine and Chelsea Finn},
      year={2019},
      eprint={1912.04226},
      archivePrefix={arXiv},
      primaryClass={cs.AI},
      url={https://arxiv.org/abs/1912.04226}, 
}

@misc{czarnecki2018mixmatchagentcurricula,
      title={Mix\&Match - Agent Curricula for Reinforcement Learning}, 
      author={Wojciech Marian Czarnecki and Siddhant M. Jayakumar and Max Jaderberg and Leonard Hasenclever and Yee Whye Teh and Simon Osindero and Nicolas Heess and Razvan Pascanu},
      year={2018},
      eprint={1806.01780},
      archivePrefix={arXiv},
      primaryClass={cs.LG},
      url={https://arxiv.org/abs/1806.01780}, 
}

@misc{rusu2022progressiveneuralnetworks,
      title={Progressive Neural Networks}, 
      author={Andrei A. Rusu and Neil C. Rabinowitz and Guillaume Desjardins and Hubert Soyer and James Kirkpatrick and Koray Kavukcuoglu and Razvan Pascanu and Raia Hadsell},
      year={2022},
      eprint={1606.04671},
      archivePrefix={arXiv},
      primaryClass={cs.LG},
      url={https://arxiv.org/abs/1606.04671}, 
}

@misc{florensa2018automaticgoalgenerationreinforcement,
      title={Automatic Goal Generation for Reinforcement Learning Agents}, 
      author={Carlos Florensa and David Held and Xinyang Geng and Pieter Abbeel},
      year={2018},
      eprint={1705.06366},
      archivePrefix={arXiv},
      primaryClass={cs.LG},
      url={https://arxiv.org/abs/1705.06366}, 
}

@misc{matiisen2017teacherstudentcurriculumlearning,
      title={Teacher-Student Curriculum Learning}, 
      author={Tambet Matiisen and Avital Oliver and Taco Cohen and John Schulman},
      year={2017},
      eprint={1707.00183},
      archivePrefix={arXiv},
      primaryClass={cs.LG},
      url={https://arxiv.org/abs/1707.00183}, 
}

@misc{sukhbaatar2018intrinsicmotivationautomaticcurricula,
      title={Intrinsic Motivation and Automatic Curricula via Asymmetric Self-Play}, 
      author={Sainbayar Sukhbaatar and Zeming Lin and Ilya Kostrikov and Gabriel Synnaeve and Arthur Szlam and Rob Fergus},
      year={2018},
      eprint={1703.05407},
      archivePrefix={arXiv},
      primaryClass={cs.LG},
      url={https://arxiv.org/abs/1703.05407}, 
}

@misc{portelas2019teacheralgorithmscurriculumlearning,
      title={Teacher algorithms for curriculum learning of Deep RL in continuously parameterized environments}, 
      author={Rémy Portelas and Cédric Colas and Katja Hofmann and Pierre-Yves Oudeyer},
      year={2019},
      eprint={1910.07224},
      archivePrefix={arXiv},
      primaryClass={cs.LG},
      url={https://arxiv.org/abs/1910.07224}, 
}

@misc{wang2019pairedopenendedtrailblazerpoet,
      title={Paired Open-Ended Trailblazer (POET): Endlessly Generating Increasingly Complex and Diverse Learning Environments and Their Solutions}, 
      author={Rui Wang and Joel Lehman and Jeff Clune and Kenneth O. Stanley},
      year={2019},
      eprint={1901.01753},
      archivePrefix={arXiv},
      primaryClass={cs.NE},
      url={https://arxiv.org/abs/1901.01753}, 
}

@misc{justesen2018illuminatinggeneralizationdeepreinforcement,
      title={Illuminating Generalization in Deep Reinforcement Learning through Procedural Level Generation}, 
      author={Niels Justesen and Ruben Rodriguez Torrado and Philip Bontrager and Ahmed Khalifa and Julian Togelius and Sebastian Risi},
      year={2018},
      eprint={1806.10729},
      archivePrefix={arXiv},
      primaryClass={cs.LG},
      url={https://arxiv.org/abs/1806.10729}, 
}

@misc{zaremba2015learningexecute,
      title={Learning to Execute}, 
      author={Wojciech Zaremba and Ilya Sutskever},
      year={2015},
      eprint={1410.4615},
      archivePrefix={arXiv},
      primaryClass={cs.NE},
      url={https://arxiv.org/abs/1410.4615}, 
}

@misc{ye2025limoreasoning,
      title={LIMO: Less is More for Reasoning}, 
      author={Yixin Ye and Zhen Huang and Yang Xiao and Ethan Chern and Shijie Xia and Pengfei Liu},
      year={2025},
      eprint={2502.03387},
      archivePrefix={arXiv},
      primaryClass={cs.CL},
      url={https://arxiv.org/abs/2502.03387}, 
}

@misc{li2025limrrlscaling,
      title={LIMR: Less is More for RL Scaling}, 
      author={Xuefeng Li and Haoyang Zou and Pengfei Liu},
      year={2025},
      eprint={2502.11886},
      archivePrefix={arXiv},
      primaryClass={cs.LG},
      url={https://arxiv.org/abs/2502.11886}, 
}

@misc{muennighoff2025s1simpletesttimescaling,
      title={s1: Simple test-time scaling}, 
      author={Niklas Muennighoff and Zitong Yang and Weijia Shi and Xiang Lisa Li and Li Fei-Fei and Hannaneh Hajishirzi and Luke Zettlemoyer and Percy Liang and Emmanuel Candès and Tatsunori Hashimoto},
      year={2025},
      eprint={2501.19393},
      archivePrefix={arXiv},
      primaryClass={cs.CL},
      url={https://arxiv.org/abs/2501.19393}, 
}

@inproceedings{bengio2009curriculumlearning,
author = {Bengio, Yoshua and Louradour, J\'{e}r\^{o}me and Collobert, Ronan and Weston, Jason},
title = {Curriculum learning},
year = {2009},
isbn = {9781605585161},
publisher = {Association for Computing Machinery},
address = {New York, NY, USA},
url = {https://doi.org/10.1145/1553374.1553380},
doi = {10.1145/1553374.1553380},
booktitle = {Proceedings of the 26th Annual International Conference on Machine Learning},
pages = {41–48},
numpages = {8},
location = {Montreal, Quebec, Canada},
series = {ICML '09}
}

@inproceedings{
lin2025zebralogic,
title={ZebraLogic: On the Scaling Limits of {LLM}s for Logical Reasoning},
author={Bill Yuchen Lin and Ronan Le Bras and Kyle Richardson and Ashish Sabharwal and Radha Poovendran and Peter Clark and Yejin Choi},
booktitle={Forty-second International Conference on Machine Learning},
year={2025},
url={https://openreview.net/forum?id=sTAJ9QyA6l}
}

@misc{hu2025reinforcesimpleefficientapproach,
      title={REINFORCE++: A Simple and Efficient Approach for Aligning Large Language Models}, 
      author={Jian Hu},
      year={2025},
      eprint={2501.03262},
      archivePrefix={arXiv},
      primaryClass={cs.CL},
      url={https://arxiv.org/abs/2501.03262}, 
}

@misc{zeng2025simplerlzooinvestigatingtamingzero,
      title={SimpleRL-Zoo: Investigating and Taming Zero Reinforcement Learning for Open Base Models in the Wild}, 
      author={Weihao Zeng and Yuzhen Huang and Qian Liu and Wei Liu and Keqing He and Zejun Ma and Junxian He},
      year={2025},
      eprint={2503.18892},
      archivePrefix={arXiv},
      primaryClass={cs.LG},
      url={https://arxiv.org/abs/2503.18892}, 
}

@misc{wang2025reinforcementlearningreasoninglarge,
      title={Reinforcement Learning for Reasoning in Large Language Models with One Training Example}, 
      author={Yiping Wang and Qing Yang and Zhiyuan Zeng and Liliang Ren and Lucas Liu and Baolin Peng and Hao Cheng and Xuehai He and Kuan Wang and Jianfeng Gao and Weizhu Chen and Shuohang Wang and Simon Shaolei Du and Yelong Shen},
      year={2025},
      eprint={2504.20571},
      archivePrefix={arXiv},
      primaryClass={cs.LG},
      url={https://arxiv.org/abs/2504.20571}, 
}

@misc{wen2025lightr1curriculumsftdpo,
      title={Light-R1: Curriculum SFT, DPO and RL for Long COT from Scratch and Beyond}, 
      author={Liang Wen and Yunke Cai and Fenrui Xiao and Xin He and Qi An and Zhenyu Duan and Yimin Du and Junchen Liu and Lifu Tang and Xiaowei Lv and Haosheng Zou and Yongchao Deng and Shousheng Jia and Xiangzheng Zhang},
      year={2025},
      eprint={2503.10460},
      archivePrefix={arXiv},
      primaryClass={cs.CL},
      url={https://arxiv.org/abs/2503.10460}, 
}

@misc{song2025fastcurlcurriculumreinforcementlearning,
      title={FastCuRL: Curriculum Reinforcement Learning with Progressive Context Extension for Efficient Training R1-like Reasoning Models}, 
      author={Mingyang Song and Mao Zheng and Zheng Li and Wenjie Yang and Xuan Luo and Yue Pan and Feng Zhang},
      year={2025},
      eprint={2503.17287},
      archivePrefix={arXiv},
      primaryClass={cs.CL},
      url={https://arxiv.org/abs/2503.17287}, 
}

@misc{hu2025openreasonerzeroopensourceapproach,
      title={Open-Reasoner-Zero: An Open Source Approach to Scaling Up Reinforcement Learning on the Base Model}, 
      author={Jingcheng Hu and Yinmin Zhang and Qi Han and Daxin Jiang and Xiangyu Zhang and Heung-Yeung Shum},
      year={2025},
      eprint={2503.24290},
      archivePrefix={arXiv},
      primaryClass={cs.LG},
      url={https://arxiv.org/abs/2503.24290}, 
}

@misc{zr1,
      title={ZR1-1.5B, a small but powerful reasoning model for math and code}, 
      author={Zyphra},
      year={2025},
      url={https://www.zyphra.com/post/introducing-zr1-1-5b-a-small-but-powerful-math-code-reasoning-model}, 
}

@misc{chen2025empiricalstudyelicitingimproving,
      title={An Empirical Study on Eliciting and Improving R1-like Reasoning Models}, 
      author={Zhipeng Chen and Yingqian Min and Beichen Zhang and Jie Chen and Jinhao Jiang and Daixuan Cheng and Wayne Xin Zhao and Zheng Liu and Xu Miao and Yang Lu and Lei Fang and Zhongyuan Wang and Ji-Rong Wen},
      year={2025},
      eprint={2503.04548},
      archivePrefix={arXiv},
      primaryClass={cs.CL},
      url={https://arxiv.org/abs/2503.04548}, 
}

@misc{yu2025dapoopensourcellmreinforcement,
      title={DAPO: An Open-Source LLM Reinforcement Learning System at Scale}, 
      author={Qiying Yu and Zheng Zhang and Ruofei Zhu and Yufeng Yuan and Xiaochen Zuo and Yu Yue and Tiantian Fan and Gaohong Liu and Lingjun Liu and Xin Liu and Haibin Lin and Zhiqi Lin and Bole Ma and Guangming Sheng and Yuxuan Tong and Chi Zhang and Mofan Zhang and Wang Zhang and Hang Zhu and Jinhua Zhu and Jiaze Chen and Jiangjie Chen and Chengyi Wang and Hongli Yu and Weinan Dai and Yuxuan Song and Xiangpeng Wei and Hao Zhou and Jingjing Liu and Wei-Ying Ma and Ya-Qin Zhang and Lin Yan and Mu Qiao and Yonghui Wu and Mingxuan Wang},
      year={2025},
      eprint={2503.14476},
      archivePrefix={arXiv},
      primaryClass={cs.LG},
      url={https://arxiv.org/abs/2503.14476}, 
}

@misc{shen2025exploringdatascalingtrends,
      title={Exploring Data Scaling Trends and Effects in Reinforcement Learning from Human Feedback}, 
      author={Wei Shen and Guanlin Liu and Zheng Wu and Ruofei Zhu and Qingping Yang and Chao Xin and Yu Yue and Lin Yan},
      year={2025},
      eprint={2503.22230},
      archivePrefix={arXiv},
      primaryClass={cs.LG},
      url={https://arxiv.org/abs/2503.22230}, 
}

@misc{cui2025processreinforcementimplicitrewards,
      title={Process Reinforcement through Implicit Rewards}, 
      author={Ganqu Cui and Lifan Yuan and Zefan Wang and Hanbin Wang and Wendi Li and Bingxiang He and Yuchen Fan and Tianyu Yu and Qixin Xu and Weize Chen and Jiarui Yuan and Huayu Chen and Kaiyan Zhang and Xingtai Lv and Shuo Wang and Yuan Yao and Xu Han and Hao Peng and Yu Cheng and Zhiyuan Liu and Maosong Sun and Bowen Zhou and Ning Ding},
      year={2025},
      eprint={2502.01456},
      archivePrefix={arXiv},
      primaryClass={cs.LG},
      url={https://arxiv.org/abs/2502.01456}, 
}

@misc{deepscaler2025,
  title={DeepScaleR: Surpassing O1-Preview with a 1.5B Model by Scaling RL},
  author={Michael Luo and Sijun Tan and Justin Wong and Xiaoxiang Shi and William Y. Tang and Manan Roongta and Colin Cai and Jeffrey Luo and Tianjun Zhang and Li Erran Li and Raluca Ada Popa and Ion Stoica},
  year={2025},
  note={Notion Blog},
  year={2025}
}

@misc{li2024remaxsimpleeffectiveefficient,
      title={ReMax: A Simple, Effective, and Efficient Reinforcement Learning Method for Aligning Large Language Models}, 
      author={Ziniu Li and Tian Xu and Yushun Zhang and Zhihang Lin and Yang Yu and Ruoyu Sun and Zhi-Quan Luo},
      year={2024},
      eprint={2310.10505},
      archivePrefix={arXiv},
      primaryClass={cs.LG},
      url={https://arxiv.org/abs/2310.10505}, 
}

@misc{lewkowycz2022solvingquantitativereasoningproblems,
      title={Solving Quantitative Reasoning Problems with Language Models}, 
      author={Aitor Lewkowycz and Anders Andreassen and David Dohan and Ethan Dyer and Henryk Michalewski and Vinay Ramasesh and Ambrose Slone and Cem Anil and Imanol Schlag and Theo Gutman-Solo and Yuhuai Wu and Behnam Neyshabur and Guy Gur-Ari and Vedant Misra},
      year={2022},
      eprint={2206.14858},
      archivePrefix={arXiv},
      primaryClass={cs.CL},
      url={https://arxiv.org/abs/2206.14858}, 
}

@misc{he2024olympiadbenchchallengingbenchmarkpromoting,
      title={OlympiadBench: A Challenging Benchmark for Promoting AGI with Olympiad-Level Bilingual Multimodal Scientific Problems}, 
      author={Chaoqun He and Renjie Luo and Yuzhuo Bai and Shengding Hu and Zhen Leng Thai and Junhao Shen and Jinyi Hu and Xu Han and Yujie Huang and Yuxiang Zhang and Jie Liu and Lei Qi and Zhiyuan Liu and Maosong Sun},
      year={2024},
      eprint={2402.14008},
      archivePrefix={arXiv},
      primaryClass={cs.CL},
      url={https://arxiv.org/abs/2402.14008}, 
}

@misc{cobbe2021trainingverifierssolvemath,
      title={Training Verifiers to Solve Math Word Problems}, 
      author={Karl Cobbe and Vineet Kosaraju and Mohammad Bavarian and Mark Chen and Heewoo Jun and Lukasz Kaiser and Matthias Plappert and Jerry Tworek and Jacob Hilton and Reiichiro Nakano and Christopher Hesse and John Schulman},
      year={2021},
      eprint={2110.14168},
      archivePrefix={arXiv},
      primaryClass={cs.LG},
      url={https://arxiv.org/abs/2110.14168}, 
}

@inproceedings{
hendrycks2021measuring,
title={Measuring Mathematical Problem Solving With the {MATH} Dataset},
author={Dan Hendrycks and Collin Burns and Saurav Kadavath and Akul Arora and Steven Basart and Eric Tang and Dawn Song and Jacob Steinhardt},
booktitle={Thirty-fifth Conference on Neural Information Processing Systems Datasets and Benchmarks Track (Round 2)},
year={2021},
url={https://openreview.net/forum?id=7Bywt2mQsCe}
}

@misc{lightman2023letsverifystepstep,
      title={Let's Verify Step by Step}, 
      author={Hunter Lightman and Vineet Kosaraju and Yura Burda and Harri Edwards and Bowen Baker and Teddy Lee and Jan Leike and John Schulman and Ilya Sutskever and Karl Cobbe},
      year={2023},
      eprint={2305.20050},
      archivePrefix={arXiv},
      primaryClass={cs.LG},
      url={https://arxiv.org/abs/2305.20050}, 
}

@misc{schulman2017proximalpolicyoptimizationalgorithms,
      title={Proximal Policy Optimization Algorithms}, 
      author={John Schulman and Filip Wolski and Prafulla Dhariwal and Alec Radford and Oleg Klimov},
      year={2017},
      eprint={1707.06347},
      archivePrefix={arXiv},
      primaryClass={cs.LG},
      url={https://arxiv.org/abs/1707.06347}, 
}

@misc{grattafiori2024llama3herdmodels,
      title={The Llama 3 Herd of Models}, 
      author={Aaron Grattafiori and Abhimanyu Dubey and Abhinav Jauhri and Abhinav Pandey and Abhishek Kadian and Ahmad Al-Dahle and Aiesha Letman and Akhil Mathur and Alan Schelten and Alex Vaughan and Amy Yang and Angela Fan and Anirudh Goyal and Anthony Hartshorn and Aobo Yang and Archi Mitra and Archie Sravankumar and Artem Korenev and Arthur Hinsvark and Arun Rao and Aston Zhang and Aurelien Rodriguez and Austen Gregerson and Ava Spataru and Baptiste Roziere and Bethany Biron and Binh Tang and Bobbie Chern and Charlotte Caucheteux and Chaya Nayak and Chloe Bi and Chris Marra and Chris McConnell and Christian Keller and Christophe Touret and Chunyang Wu and Corinne Wong and Cristian Canton Ferrer and Cyrus Nikolaidis and Damien Allonsius and Daniel Song and Danielle Pintz and Danny Livshits and Danny Wyatt and David Esiobu and Dhruv Choudhary and Dhruv Mahajan and Diego Garcia-Olano and Diego Perino and Dieuwke Hupkes and Egor Lakomkin and Ehab AlBadawy and Elina Lobanova and Emily Dinan and Eric Michael Smith and Filip Radenovic and Francisco Guzmán and Frank Zhang and Gabriel Synnaeve and Gabrielle Lee and Georgia Lewis Anderson and Govind Thattai and Graeme Nail and Gregoire Mialon and Guan Pang and Guillem Cucurell and Hailey Nguyen and Hannah Korevaar and Hu Xu and Hugo Touvron and Iliyan Zarov and Imanol Arrieta Ibarra and Isabel Kloumann and Ishan Misra and Ivan Evtimov and Jack Zhang and Jade Copet and Jaewon Lee and Jan Geffert and Jana Vranes and Jason Park and Jay Mahadeokar and Jeet Shah and Jelmer van der Linde and Jennifer Billock and Jenny Hong and Jenya Lee and Jeremy Fu and Jianfeng Chi and Jianyu Huang and Jiawen Liu and Jie Wang and Jiecao Yu and Joanna Bitton and Joe Spisak and Jongsoo Park and Joseph Rocca and Joshua Johnstun and Joshua Saxe and Junteng Jia and Kalyan Vasuden Alwala and Karthik Prasad and Kartikeya Upasani and Kate Plawiak and Ke Li and Kenneth Heafield and Kevin Stone and Khalid El-Arini and Krithika Iyer and Kshitiz Malik and Kuenley Chiu and Kunal Bhalla and Kushal Lakhotia and Lauren Rantala-Yeary and Laurens van der Maaten and Lawrence Chen and Liang Tan and Liz Jenkins and Louis Martin and Lovish Madaan and Lubo Malo and Lukas Blecher and Lukas Landzaat and Luke de Oliveira and Madeline Muzzi and Mahesh Pasupuleti and Mannat Singh and Manohar Paluri and Marcin Kardas and Maria Tsimpoukelli and Mathew Oldham and Mathieu Rita and Maya Pavlova and Melanie Kambadur and Mike Lewis and Min Si and Mitesh Kumar Singh and Mona Hassan and Naman Goyal and Narjes Torabi and Nikolay Bashlykov and Nikolay Bogoychev and Niladri Chatterji and Ning Zhang and Olivier Duchenne and Onur Çelebi and Patrick Alrassy and Pengchuan Zhang and Pengwei Li and Petar Vasic and Peter Weng and Prajjwal Bhargava and Pratik Dubal and Praveen Krishnan and Punit Singh Koura and Puxin Xu and Qing He and Qingxiao Dong and Ragavan Srinivasan and Raj Ganapathy and Ramon Calderer and Ricardo Silveira Cabral and Robert Stojnic and Roberta Raileanu and Rohan Maheswari and Rohit Girdhar and Rohit Patel and Romain Sauvestre and Ronnie Polidoro and Roshan Sumbaly and Ross Taylor and Ruan Silva and Rui Hou and Rui Wang and Saghar Hosseini and Sahana Chennabasappa and Sanjay Singh and Sean Bell and Seohyun Sonia Kim and Sergey Edunov and Shaoliang Nie and Sharan Narang and Sharath Raparthy and Sheng Shen and Shengye Wan and Shruti Bhosale and Shun Zhang and Simon Vandenhende and Soumya Batra and Spencer Whitman and Sten Sootla and Stephane Collot and Suchin Gururangan and Sydney Borodinsky and Tamar Herman and Tara Fowler and Tarek Sheasha and Thomas Georgiou and Thomas Scialom and Tobias Speckbacher and Todor Mihaylov and Tong Xiao and Ujjwal Karn and Vedanuj Goswami and Vibhor Gupta and Vignesh Ramanathan and Viktor Kerkez and Vincent Gonguet and Virginie Do and Vish Vogeti and Vítor Albiero and Vladan Petrovic and Weiwei Chu and Wenhan Xiong and Wenyin Fu and Whitney Meers and Xavier Martinet and Xiaodong Wang and Xiaofang Wang and Xiaoqing Ellen Tan and Xide Xia and Xinfeng Xie and Xuchao Jia and Xuewei Wang and Yaelle Goldschlag and Yashesh Gaur and Yasmine Babaei and Yi Wen and Yiwen Song and Yuchen Zhang and Yue Li and Yuning Mao and Zacharie Delpierre Coudert and Zheng Yan and Zhengxing Chen and Zoe Papakipos and Aaditya Singh and Aayushi Srivastava and Abha Jain and Adam Kelsey and Adam Shajnfeld and Adithya Gangidi and Adolfo Victoria and Ahuva Goldstand and Ajay Menon and Ajay Sharma and Alex Boesenberg and Alexei Baevski and Allie Feinstein and Amanda Kallet and Amit Sangani and Amos Teo and Anam Yunus and Andrei Lupu and Andres Alvarado and Andrew Caples and Andrew Gu and Andrew Ho and Andrew Poulton and Andrew Ryan and Ankit Ramchandani and Annie Dong and Annie Franco and Anuj Goyal and Aparajita Saraf and Arkabandhu Chowdhury and Ashley Gabriel and Ashwin Bharambe and Assaf Eisenman and Azadeh Yazdan and Beau James and Ben Maurer and Benjamin Leonhardi and Bernie Huang and Beth Loyd and Beto De Paola and Bhargavi Paranjape and Bing Liu and Bo Wu and Boyu Ni and Braden Hancock and Bram Wasti and Brandon Spence and Brani Stojkovic and Brian Gamido and Britt Montalvo and Carl Parker and Carly Burton and Catalina Mejia and Ce Liu and Changhan Wang and Changkyu Kim and Chao Zhou and Chester Hu and Ching-Hsiang Chu and Chris Cai and Chris Tindal and Christoph Feichtenhofer and Cynthia Gao and Damon Civin and Dana Beaty and Daniel Kreymer and Daniel Li and David Adkins and David Xu and Davide Testuggine and Delia David and Devi Parikh and Diana Liskovich and Didem Foss and Dingkang Wang and Duc Le and Dustin Holland and Edward Dowling and Eissa Jamil and Elaine Montgomery and Eleonora Presani and Emily Hahn and Emily Wood and Eric-Tuan Le and Erik Brinkman and Esteban Arcaute and Evan Dunbar and Evan Smothers and Fei Sun and Felix Kreuk and Feng Tian and Filippos Kokkinos and Firat Ozgenel and Francesco Caggioni and Frank Kanayet and Frank Seide and Gabriela Medina Florez and Gabriella Schwarz and Gada Badeer and Georgia Swee and Gil Halpern and Grant Herman and Grigory Sizov and Guangyi and Zhang and Guna Lakshminarayanan and Hakan Inan and Hamid Shojanazeri and Han Zou and Hannah Wang and Hanwen Zha and Haroun Habeeb and Harrison Rudolph and Helen Suk and Henry Aspegren and Hunter Goldman and Hongyuan Zhan and Ibrahim Damlaj and Igor Molybog and Igor Tufanov and Ilias Leontiadis and Irina-Elena Veliche and Itai Gat and Jake Weissman and James Geboski and James Kohli and Janice Lam and Japhet Asher and Jean-Baptiste Gaya and Jeff Marcus and Jeff Tang and Jennifer Chan and Jenny Zhen and Jeremy Reizenstein and Jeremy Teboul and Jessica Zhong and Jian Jin and Jingyi Yang and Joe Cummings and Jon Carvill and Jon Shepard and Jonathan McPhie and Jonathan Torres and Josh Ginsburg and Junjie Wang and Kai Wu and Kam Hou U and Karan Saxena and Kartikay Khandelwal and Katayoun Zand and Kathy Matosich and Kaushik Veeraraghavan and Kelly Michelena and Keqian Li and Kiran Jagadeesh and Kun Huang and Kunal Chawla and Kyle Huang and Lailin Chen and Lakshya Garg and Lavender A and Leandro Silva and Lee Bell and Lei Zhang and Liangpeng Guo and Licheng Yu and Liron Moshkovich and Luca Wehrstedt and Madian Khabsa and Manav Avalani and Manish Bhatt and Martynas Mankus and Matan Hasson and Matthew Lennie and Matthias Reso and Maxim Groshev and Maxim Naumov and Maya Lathi and Meghan Keneally and Miao Liu and Michael L. Seltzer and Michal Valko and Michelle Restrepo and Mihir Patel and Mik Vyatskov and Mikayel Samvelyan and Mike Clark and Mike Macey and Mike Wang and Miquel Jubert Hermoso and Mo Metanat and Mohammad Rastegari and Munish Bansal and Nandhini Santhanam and Natascha Parks and Natasha White and Navyata Bawa and Nayan Singhal and Nick Egebo and Nicolas Usunier and Nikhil Mehta and Nikolay Pavlovich Laptev and Ning Dong and Norman Cheng and Oleg Chernoguz and Olivia Hart and Omkar Salpekar and Ozlem Kalinli and Parkin Kent and Parth Parekh and Paul Saab and Pavan Balaji and Pedro Rittner and Philip Bontrager and Pierre Roux and Piotr Dollar and Polina Zvyagina and Prashant Ratanchandani and Pritish Yuvraj and Qian Liang and Rachad Alao and Rachel Rodriguez and Rafi Ayub and Raghotham Murthy and Raghu Nayani and Rahul Mitra and Rangaprabhu Parthasarathy and Raymond Li and Rebekkah Hogan and Robin Battey and Rocky Wang and Russ Howes and Ruty Rinott and Sachin Mehta and Sachin Siby and Sai Jayesh Bondu and Samyak Datta and Sara Chugh and Sara Hunt and Sargun Dhillon and Sasha Sidorov and Satadru Pan and Saurabh Mahajan and Saurabh Verma and Seiji Yamamoto and Sharadh Ramaswamy and Shaun Lindsay and Shaun Lindsay and Sheng Feng and Shenghao Lin and Shengxin Cindy Zha and Shishir Patil and Shiva Shankar and Shuqiang Zhang and Shuqiang Zhang and Sinong Wang and Sneha Agarwal and Soji Sajuyigbe and Soumith Chintala and Stephanie Max and Stephen Chen and Steve Kehoe and Steve Satterfield and Sudarshan Govindaprasad and Sumit Gupta and Summer Deng and Sungmin Cho and Sunny Virk and Suraj Subramanian and Sy Choudhury and Sydney Goldman and Tal Remez and Tamar Glaser and Tamara Best and Thilo Koehler and Thomas Robinson and Tianhe Li and Tianjun Zhang and Tim Matthews and Timothy Chou and Tzook Shaked and Varun Vontimitta and Victoria Ajayi and Victoria Montanez and Vijai Mohan and Vinay Satish Kumar and Vishal Mangla and Vlad Ionescu and Vlad Poenaru and Vlad Tiberiu Mihailescu and Vladimir Ivanov and Wei Li and Wenchen Wang and Wenwen Jiang and Wes Bouaziz and Will Constable and Xiaocheng Tang and Xiaojian Wu and Xiaolan Wang and Xilun Wu and Xinbo Gao and Yaniv Kleinman and Yanjun Chen and Ye Hu and Ye Jia and Ye Qi and Yenda Li and Yilin Zhang and Ying Zhang and Yossi Adi and Youngjin Nam and Yu and Wang and Yu Zhao and Yuchen Hao and Yundi Qian and Yunlu Li and Yuzi He and Zach Rait and Zachary DeVito and Zef Rosnbrick and Zhaoduo Wen and Zhenyu Yang and Zhiwei Zhao and Zhiyu Ma},
      year={2024},
      eprint={2407.21783},
      archivePrefix={arXiv},
      primaryClass={cs.AI},
      url={https://arxiv.org/abs/2407.21783}, 
}

@article{sheng2024hybridflow,
  title   = {HybridFlow: A Flexible and Efficient RLHF Framework},
  author  = {Guangming Sheng and Chi Zhang and Zilingfeng Ye and Xibin Wu and Wang Zhang and Ru Zhang and Yanghua Peng and Haibin Lin and Chuan Wu},
  year    = {2024},
  journal = {arXiv preprint arXiv: 2409.19256}
}

@misc{ahmadian2024basicsrevisitingreinforcestyle,
      title={Back to Basics: Revisiting REINFORCE Style Optimization for Learning from Human Feedback in LLMs}, 
      author={Arash Ahmadian and Chris Cremer and Matthias Gallé and Marzieh Fadaee and Julia Kreutzer and Olivier Pietquin and Ahmet Üstün and Sara Hooker},
      year={2024},
      eprint={2402.14740},
      archivePrefix={arXiv},
      primaryClass={cs.LG},
      url={https://arxiv.org/abs/2402.14740}, 
}

@misc{deepseekai2025deepseekr1incentivizingreasoningcapability,
      title={DeepSeek-R1: Incentivizing Reasoning Capability in LLMs via Reinforcement Learning}, 
      author={DeepSeek-AI and Daya Guo and Dejian Yang and Haowei Zhang and Junxiao Song and Ruoyu Zhang and Runxin Xu and Qihao Zhu and Shirong Ma and Peiyi Wang and Xiao Bi and Xiaokang Zhang and Xingkai Yu and Yu Wu and Z. F. Wu and Zhibin Gou and Zhihong Shao and Zhuoshu Li and Ziyi Gao and Aixin Liu and Bing Xue and Bingxuan Wang and Bochao Wu and Bei Feng and Chengda Lu and Chenggang Zhao and Chengqi Deng and Chenyu Zhang and Chong Ruan and Damai Dai and Deli Chen and Dongjie Ji and Erhang Li and Fangyun Lin and Fucong Dai and Fuli Luo and Guangbo Hao and Guanting Chen and Guowei Li and H. Zhang and Han Bao and Hanwei Xu and Haocheng Wang and Honghui Ding and Huajian Xin and Huazuo Gao and Hui Qu and Hui Li and Jianzhong Guo and Jiashi Li and Jiawei Wang and Jingchang Chen and Jingyang Yuan and Junjie Qiu and Junlong Li and J. L. Cai and Jiaqi Ni and Jian Liang and Jin Chen and Kai Dong and Kai Hu and Kaige Gao and Kang Guan and Kexin Huang and Kuai Yu and Lean Wang and Lecong Zhang and Liang Zhao and Litong Wang and Liyue Zhang and Lei Xu and Leyi Xia and Mingchuan Zhang and Minghua Zhang and Minghui Tang and Meng Li and Miaojun Wang and Mingming Li and Ning Tian and Panpan Huang and Peng Zhang and Qiancheng Wang and Qinyu Chen and Qiushi Du and Ruiqi Ge and Ruisong Zhang and Ruizhe Pan and Runji Wang and R. J. Chen and R. L. Jin and Ruyi Chen and Shanghao Lu and Shangyan Zhou and Shanhuang Chen and Shengfeng Ye and Shiyu Wang and Shuiping Yu and Shunfeng Zhou and Shuting Pan and S. S. Li and Shuang Zhou and Shaoqing Wu and Shengfeng Ye and Tao Yun and Tian Pei and Tianyu Sun and T. Wang and Wangding Zeng and Wanjia Zhao and Wen Liu and Wenfeng Liang and Wenjun Gao and Wenqin Yu and Wentao Zhang and W. L. Xiao and Wei An and Xiaodong Liu and Xiaohan Wang and Xiaokang Chen and Xiaotao Nie and Xin Cheng and Xin Liu and Xin Xie and Xingchao Liu and Xinyu Yang and Xinyuan Li and Xuecheng Su and Xuheng Lin and X. Q. Li and Xiangyue Jin and Xiaojin Shen and Xiaosha Chen and Xiaowen Sun and Xiaoxiang Wang and Xinnan Song and Xinyi Zhou and Xianzu Wang and Xinxia Shan and Y. K. Li and Y. Q. Wang and Y. X. Wei and Yang Zhang and Yanhong Xu and Yao Li and Yao Zhao and Yaofeng Sun and Yaohui Wang and Yi Yu and Yichao Zhang and Yifan Shi and Yiliang Xiong and Ying He and Yishi Piao and Yisong Wang and Yixuan Tan and Yiyang Ma and Yiyuan Liu and Yongqiang Guo and Yuan Ou and Yuduan Wang and Yue Gong and Yuheng Zou and Yujia He and Yunfan Xiong and Yuxiang Luo and Yuxiang You and Yuxuan Liu and Yuyang Zhou and Y. X. Zhu and Yanhong Xu and Yanping Huang and Yaohui Li and Yi Zheng and Yuchen Zhu and Yunxian Ma and Ying Tang and Yukun Zha and Yuting Yan and Z. Z. Ren and Zehui Ren and Zhangli Sha and Zhe Fu and Zhean Xu and Zhenda Xie and Zhengyan Zhang and Zhewen Hao and Zhicheng Ma and Zhigang Yan and Zhiyu Wu and Zihui Gu and Zijia Zhu and Zijun Liu and Zilin Li and Ziwei Xie and Ziyang Song and Zizheng Pan and Zhen Huang and Zhipeng Xu and Zhongyu Zhang and Zhen Zhang},
      year={2025},
      eprint={2501.12948},
      archivePrefix={arXiv},
      primaryClass={cs.CL},
      url={https://arxiv.org/abs/2501.12948}, 
}

@article{bae2025online,
  title={Online Difficulty Filtering for Reasoning Oriented Reinforcement Learning},
  author={Bae, Sanghwan and Hong, Jiwoo and Lee, Min Young and Kim, Hanbyul and Nam, JeongYeon and Kwak, Donghyun},
  journal={arXiv preprint arXiv:2504.03380},
  year={2025}
}

@article{rafailov2024r,
  title={From r to q*: Your language model is secretly a q-function},
  author={Rafailov, Rafael and Hejna, Joey and Park, Ryan and Finn, Chelsea},
  journal={arXiv preprint arXiv:2404.12358},
  year={2024}
}

@inproceedings{haarnoja2017reinforcement,
  title={Reinforcement learning with deep energy-based policies},
  author={Haarnoja, Tuomas and Tang, Haoran and Abbeel, Pieter and Levine, Sergey},
  booktitle={International conference on machine learning},
  pages={1352--1361},
  year={2017},
  organization={PMLR}
}

@misc{openai2024gpt4ocard,
      title={GPT-4o System Card}, 
      author={OpenAI and : and Aaron Hurst and Adam Lerer and Adam P. Goucher and Adam Perelman and Aditya Ramesh and Aidan Clark and AJ Ostrow and Akila Welihinda and Alan Hayes and Alec Radford and Aleksander Mądry and Alex Baker-Whitcomb and Alex Beutel and Alex Borzunov and Alex Carney and Alex Chow and Alex Kirillov and Alex Nichol and Alex Paino and Alex Renzin and Alex Tachard Passos and Alexander Kirillov and Alexi Christakis and Alexis Conneau and Ali Kamali and Allan Jabri and Allison Moyer and Allison Tam and Amadou Crookes and Amin Tootoochian and Amin Tootoonchian and Ananya Kumar and Andrea Vallone and Andrej Karpathy and Andrew Braunstein and Andrew Cann and Andrew Codispoti and Andrew Galu and Andrew Kondrich and Andrew Tulloch and Andrey Mishchenko and Angela Baek and Angela Jiang and Antoine Pelisse and Antonia Woodford and Anuj Gosalia and Arka Dhar and Ashley Pantuliano and Avi Nayak and Avital Oliver and Barret Zoph and Behrooz Ghorbani and Ben Leimberger and Ben Rossen and Ben Sokolowsky and Ben Wang and Benjamin Zweig and Beth Hoover and Blake Samic and Bob McGrew and Bobby Spero and Bogo Giertler and Bowen Cheng and Brad Lightcap and Brandon Walkin and Brendan Quinn and Brian Guarraci and Brian Hsu and Bright Kellogg and Brydon Eastman and Camillo Lugaresi and Carroll Wainwright and Cary Bassin and Cary Hudson and Casey Chu and Chad Nelson and Chak Li and Chan Jun Shern and Channing Conger and Charlotte Barette and Chelsea Voss and Chen Ding and Cheng Lu and Chong Zhang and Chris Beaumont and Chris Hallacy and Chris Koch and Christian Gibson and Christina Kim and Christine Choi and Christine McLeavey and Christopher Hesse and Claudia Fischer and Clemens Winter and Coley Czarnecki and Colin Jarvis and Colin Wei and Constantin Koumouzelis and Dane Sherburn and Daniel Kappler and Daniel Levin and Daniel Levy and David Carr and David Farhi and David Mely and David Robinson and David Sasaki and Denny Jin and Dev Valladares and Dimitris Tsipras and Doug Li and Duc Phong Nguyen and Duncan Findlay and Edede Oiwoh and Edmund Wong and Ehsan Asdar and Elizabeth Proehl and Elizabeth Yang and Eric Antonow and Eric Kramer and Eric Peterson and Eric Sigler and Eric Wallace and Eugene Brevdo and Evan Mays and Farzad Khorasani and Felipe Petroski Such and Filippo Raso and Francis Zhang and Fred von Lohmann and Freddie Sulit and Gabriel Goh and Gene Oden and Geoff Salmon and Giulio Starace and Greg Brockman and Hadi Salman and Haiming Bao and Haitang Hu and Hannah Wong and Haoyu Wang and Heather Schmidt and Heather Whitney and Heewoo Jun and Hendrik Kirchner and Henrique Ponde de Oliveira Pinto and Hongyu Ren and Huiwen Chang and Hyung Won Chung and Ian Kivlichan and Ian O'Connell and Ian O'Connell and Ian Osband and Ian Silber and Ian Sohl and Ibrahim Okuyucu and Ikai Lan and Ilya Kostrikov and Ilya Sutskever and Ingmar Kanitscheider and Ishaan Gulrajani and Jacob Coxon and Jacob Menick and Jakub Pachocki and James Aung and James Betker and James Crooks and James Lennon and Jamie Kiros and Jan Leike and Jane Park and Jason Kwon and Jason Phang and Jason Teplitz and Jason Wei and Jason Wolfe and Jay Chen and Jeff Harris and Jenia Varavva and Jessica Gan Lee and Jessica Shieh and Ji Lin and Jiahui Yu and Jiayi Weng and Jie Tang and Jieqi Yu and Joanne Jang and Joaquin Quinonero Candela and Joe Beutler and Joe Landers and Joel Parish and Johannes Heidecke and John Schulman and Jonathan Lachman and Jonathan McKay and Jonathan Uesato and Jonathan Ward and Jong Wook Kim and Joost Huizinga and Jordan Sitkin and Jos Kraaijeveld and Josh Gross and Josh Kaplan and Josh Snyder and Joshua Achiam and Joy Jiao and Joyce Lee and Juntang Zhuang and Justyn Harriman and Kai Fricke and Kai Hayashi and Karan Singhal and Katy Shi and Kavin Karthik and Kayla Wood and Kendra Rimbach and Kenny Hsu and Kenny Nguyen and Keren Gu-Lemberg and Kevin Button and Kevin Liu and Kiel Howe and Krithika Muthukumar and Kyle Luther and Lama Ahmad and Larry Kai and Lauren Itow and Lauren Workman and Leher Pathak and Leo Chen and Li Jing and Lia Guy and Liam Fedus and Liang Zhou and Lien Mamitsuka and Lilian Weng and Lindsay McCallum and Lindsey Held and Long Ouyang and Louis Feuvrier and Lu Zhang and Lukas Kondraciuk and Lukasz Kaiser and Luke Hewitt and Luke Metz and Lyric Doshi and Mada Aflak and Maddie Simens and Madelaine Boyd and Madeleine Thompson and Marat Dukhan and Mark Chen and Mark Gray and Mark Hudnall and Marvin Zhang and Marwan Aljubeh and Mateusz Litwin and Matthew Zeng and Max Johnson and Maya Shetty and Mayank Gupta and Meghan Shah and Mehmet Yatbaz and Meng Jia Yang and Mengchao Zhong and Mia Glaese and Mianna Chen and Michael Janner and Michael Lampe and Michael Petrov and Michael Wu and Michele Wang and Michelle Fradin and Michelle Pokrass and Miguel Castro and Miguel Oom Temudo de Castro and Mikhail Pavlov and Miles Brundage and Miles Wang and Minal Khan and Mira Murati and Mo Bavarian and Molly Lin and Murat Yesildal and Nacho Soto and Natalia Gimelshein and Natalie Cone and Natalie Staudacher and Natalie Summers and Natan LaFontaine and Neil Chowdhury and Nick Ryder and Nick Stathas and Nick Turley and Nik Tezak and Niko Felix and Nithanth Kudige and Nitish Keskar and Noah Deutsch and Noel Bundick and Nora Puckett and Ofir Nachum and Ola Okelola and Oleg Boiko and Oleg Murk and Oliver Jaffe and Olivia Watkins and Olivier Godement and Owen Campbell-Moore and Patrick Chao and Paul McMillan and Pavel Belov and Peng Su and Peter Bak and Peter Bakkum and Peter Deng and Peter Dolan and Peter Hoeschele and Peter Welinder and Phil Tillet and Philip Pronin and Philippe Tillet and Prafulla Dhariwal and Qiming Yuan and Rachel Dias and Rachel Lim and Rahul Arora and Rajan Troll and Randall Lin and Rapha Gontijo Lopes and Raul Puri and Reah Miyara and Reimar Leike and Renaud Gaubert and Reza Zamani and Ricky Wang and Rob Donnelly and Rob Honsby and Rocky Smith and Rohan Sahai and Rohit Ramchandani and Romain Huet and Rory Carmichael and Rowan Zellers and Roy Chen and Ruby Chen and Ruslan Nigmatullin and Ryan Cheu and Saachi Jain and Sam Altman and Sam Schoenholz and Sam Toizer and Samuel Miserendino and Sandhini Agarwal and Sara Culver and Scott Ethersmith and Scott Gray and Sean Grove and Sean Metzger and Shamez Hermani and Shantanu Jain and Shengjia Zhao and Sherwin Wu and Shino Jomoto and Shirong Wu and Shuaiqi and Xia and Sonia Phene and Spencer Papay and Srinivas Narayanan and Steve Coffey and Steve Lee and Stewart Hall and Suchir Balaji and Tal Broda and Tal Stramer and Tao Xu and Tarun Gogineni and Taya Christianson and Ted Sanders and Tejal Patwardhan and Thomas Cunninghman and Thomas Degry and Thomas Dimson and Thomas Raoux and Thomas Shadwell and Tianhao Zheng and Todd Underwood and Todor Markov and Toki Sherbakov and Tom Rubin and Tom Stasi and Tomer Kaftan and Tristan Heywood and Troy Peterson and Tyce Walters and Tyna Eloundou and Valerie Qi and Veit Moeller and Vinnie Monaco and Vishal Kuo and Vlad Fomenko and Wayne Chang and Weiyi Zheng and Wenda Zhou and Wesam Manassra and Will Sheu and Wojciech Zaremba and Yash Patil and Yilei Qian and Yongjik Kim and Youlong Cheng and Yu Zhang and Yuchen He and Yuchen Zhang and Yujia Jin and Yunxing Dai and Yury Malkov},
      year={2024},
      eprint={2410.21276},
      archivePrefix={arXiv},
      primaryClass={cs.CL},
      url={https://arxiv.org/abs/2410.21276}, 
}

@article{richemond2024offline,
  title={Offline regularised reinforcement learning for large language models alignment},
  author={Richemond, Pierre Harvey and Tang, Yunhao and Guo, Daniel and Calandriello, Daniele and Azar, Mohammad Gheshlaghi and Rafailov, Rafael and Pires, Bernardo Avila and Tarassov, Eugene and Spangher, Lucas and Ellsworth, Will and others},
  journal={arXiv preprint arXiv:2405.19107},
  year={2024}
}

@article{schulman2017equivalence,
  title={Equivalence between policy gradients and soft q-learning},
  author={Schulman, John and Chen, Xi and Abbeel, Pieter},
  journal={arXiv preprint arXiv:1704.06440},
  year={2017}
}

@article{go2023aligning,
  title={Aligning language models with preferences through f-divergence minimization},
  author={Go, Dongyoung and Korbak, Tomasz and Kruszewski, Germ{\'a}n and Rozen, Jos and Ryu, Nahyeon and Dymetman, Marc},
  journal={arXiv preprint arXiv:2302.08215},
  year={2023}
}

@article{rafailov2023direct,
  title={Direct preference optimization: Your language model is secretly a reward model},
  author={Rafailov, Rafael and Sharma, Archit and Mitchell, Eric and Manning, Christopher D and Ermon, Stefano and Finn, Chelsea},
  journal={Advances in Neural Information Processing Systems},
  volume={36},
  pages={53728--53741},
  year={2023}
}

@article{korbak2022reinforcement,
  title={On reinforcement learning and distribution matching for fine-tuning language models with no catastrophic forgetting},
  author={Korbak, Tomasz and Elsahar, Hady and Kruszewski, Germ{\'a}n and Dymetman, Marc},
  journal={Advances in Neural Information Processing Systems},
  volume={35},
  pages={16203--16220},
  year={2022}
}

@misc{openai2024openaio1card,
      title={OpenAI o1 System Card}, 
      author={OpenAI and : and Aaron Jaech and Adam Kalai and Adam Lerer and Adam Richardson and Ahmed El-Kishky and Aiden Low and Alec Helyar and Aleksander Madry and Alex Beutel and Alex Carney and Alex Iftimie and Alex Karpenko and Alex Tachard Passos and Alexander Neitz and Alexander Prokofiev and Alexander Wei and Allison Tam and Ally Bennett and Ananya Kumar and Andre Saraiva and Andrea Vallone and Andrew Duberstein and Andrew Kondrich and Andrey Mishchenko and Andy Applebaum and Angela Jiang and Ashvin Nair and Barret Zoph and Behrooz Ghorbani and Ben Rossen and Benjamin Sokolowsky and Boaz Barak and Bob McGrew and Borys Minaiev and Botao Hao and Bowen Baker and Brandon Houghton and Brandon McKinzie and Brydon Eastman and Camillo Lugaresi and Cary Bassin and Cary Hudson and Chak Ming Li and Charles de Bourcy and Chelsea Voss and Chen Shen and Chong Zhang and Chris Koch and Chris Orsinger and Christopher Hesse and Claudia Fischer and Clive Chan and Dan Roberts and Daniel Kappler and Daniel Levy and Daniel Selsam and David Dohan and David Farhi and David Mely and David Robinson and Dimitris Tsipras and Doug Li and Dragos Oprica and Eben Freeman and Eddie Zhang and Edmund Wong and Elizabeth Proehl and Enoch Cheung and Eric Mitchell and Eric Wallace and Erik Ritter and Evan Mays and Fan Wang and Felipe Petroski Such and Filippo Raso and Florencia Leoni and Foivos Tsimpourlas and Francis Song and Fred von Lohmann and Freddie Sulit and Geoff Salmon and Giambattista Parascandolo and Gildas Chabot and Grace Zhao and Greg Brockman and Guillaume Leclerc and Hadi Salman and Haiming Bao and Hao Sheng and Hart Andrin and Hessam Bagherinezhad and Hongyu Ren and Hunter Lightman and Hyung Won Chung and Ian Kivlichan and Ian O'Connell and Ian Osband and Ignasi Clavera Gilaberte and Ilge Akkaya and Ilya Kostrikov and Ilya Sutskever and Irina Kofman and Jakub Pachocki and James Lennon and Jason Wei and Jean Harb and Jerry Twore and Jiacheng Feng and Jiahui Yu and Jiayi Weng and Jie Tang and Jieqi Yu and Joaquin Quiñonero Candela and Joe Palermo and Joel Parish and Johannes Heidecke and John Hallman and John Rizzo and Jonathan Gordon and Jonathan Uesato and Jonathan Ward and Joost Huizinga and Julie Wang and Kai Chen and Kai Xiao and Karan Singhal and Karina Nguyen and Karl Cobbe and Katy Shi and Kayla Wood and Kendra Rimbach and Keren Gu-Lemberg and Kevin Liu and Kevin Lu and Kevin Stone and Kevin Yu and Lama Ahmad and Lauren Yang and Leo Liu and Leon Maksin and Leyton Ho and Liam Fedus and Lilian Weng and Linden Li and Lindsay McCallum and Lindsey Held and Lorenz Kuhn and Lukas Kondraciuk and Lukasz Kaiser and Luke Metz and Madelaine Boyd and Maja Trebacz and Manas Joglekar and Mark Chen and Marko Tintor and Mason Meyer and Matt Jones and Matt Kaufer and Max Schwarzer and Meghan Shah and Mehmet Yatbaz and Melody Y. Guan and Mengyuan Xu and Mengyuan Yan and Mia Glaese and Mianna Chen and Michael Lampe and Michael Malek and Michele Wang and Michelle Fradin and Mike McClay and Mikhail Pavlov and Miles Wang and Mingxuan Wang and Mira Murati and Mo Bavarian and Mostafa Rohaninejad and Nat McAleese and Neil Chowdhury and Neil Chowdhury and Nick Ryder and Nikolas Tezak and Noam Brown and Ofir Nachum and Oleg Boiko and Oleg Murk and Olivia Watkins and Patrick Chao and Paul Ashbourne and Pavel Izmailov and Peter Zhokhov and Rachel Dias and Rahul Arora and Randall Lin and Rapha Gontijo Lopes and Raz Gaon and Reah Miyara and Reimar Leike and Renny Hwang and Rhythm Garg and Robin Brown and Roshan James and Rui Shu and Ryan Cheu and Ryan Greene and Saachi Jain and Sam Altman and Sam Toizer and Sam Toyer and Samuel Miserendino and Sandhini Agarwal and Santiago Hernandez and Sasha Baker and Scott McKinney and Scottie Yan and Shengjia Zhao and Shengli Hu and Shibani Santurkar and Shraman Ray Chaudhuri and Shuyuan Zhang and Siyuan Fu and Spencer Papay and Steph Lin and Suchir Balaji and Suvansh Sanjeev and Szymon Sidor and Tal Broda and Aidan Clark and Tao Wang and Taylor Gordon and Ted Sanders and Tejal Patwardhan and Thibault Sottiaux and Thomas Degry and Thomas Dimson and Tianhao Zheng and Timur Garipov and Tom Stasi and Trapit Bansal and Trevor Creech and Troy Peterson and Tyna Eloundou and Valerie Qi and Vineet Kosaraju and Vinnie Monaco and Vitchyr Pong and Vlad Fomenko and Weiyi Zheng and Wenda Zhou and Wes McCabe and Wojciech Zaremba and Yann Dubois and Yinghai Lu and Yining Chen and Young Cha and Yu Bai and Yuchen He and Yuchen Zhang and Yunyun Wang and Zheng Shao and Zhuohan Li},
      year={2024},
      eprint={2412.16720},
      archivePrefix={arXiv},
      primaryClass={cs.AI},
      url={https://arxiv.org/abs/2412.16720}, 
}

@misc{parashar2025curriculumreinforcementlearningeasy,
      title={Curriculum Reinforcement Learning from Easy to Hard Tasks Improves LLM Reasoning}, 
      author={Shubham Parashar and Shurui Gui and Xiner Li and Hongyi Ling and Sushil Vemuri and Blake Olson and Eric Li and Yu Zhang and James Caverlee and Dileep Kalathil and Shuiwang Ji},
      year={2025},
      eprint={2506.06632},
      archivePrefix={arXiv},
      primaryClass={cs.LG},
      url={https://arxiv.org/abs/2506.06632}, 
}

@misc{kimiteam2025kimik15scalingreinforcement,
      title={Kimi k1.5: Scaling Reinforcement Learning with LLMs}, 
      author={Kimi Kimi and Angang Du and Bofei Gao and Bowei Xing and Changjiu Jiang and Cheng Chen and Cheng Li and Chenjun Xiao and Chenzhuang Du and Chonghua Liao and Chuning Tang and Congcong Wang and Dehao Zhang and Enming Yuan and Enzhe Lu and Fengxiang Tang and Flood Sung and Guangda Wei and Guokun Lai and Haiqing Guo and Han Zhu and Hao Ding and Hao Hu and Hao Yang and Hao Zhang and Haotian Yao and Haotian Zhao and Haoyu Lu and Haoze Li and Haozhen Yu and Hongcheng Gao and Huabin Zheng and Huan Yuan and Jia Chen and Jianhang Guo and Jianlin Su and Jianzhou Wang and Jie Zhao and Jin Zhang and Jingyuan Liu and Junjie Yan and Junyan Wu and Lidong Shi and Ling Ye and Longhui Yu and Mengnan Dong and Neo Zhang and Ningchen Ma and Qiwei Pan and Qucheng Gong and Shaowei Liu and Shengling Ma and Shupeng Wei and Sihan Cao and Siying Huang and Tao Jiang and Weihao Gao and Weimin Xiong and Weiran He and Weixiao Huang and Weixin Xu and Wenhao Wu and Wenyang He and Xianghui Wei and Xianqing Jia and Xingzhe Wu and Xinran Xu and Xinxing Zu and Xinyu Zhou and Xuehai Pan and Y. Charles and Yang Li and Yangyang Hu and Yangyang Liu and Yanru Chen and Yejie Wang and Yibo Liu and Yidao Qin and Yifeng Liu and Ying Yang and Yiping Bao and Yulun Du and Yuxin Wu and Yuzhi Wang and Zaida Zhou and Zhaoji Wang and Zhaowei Li and Zhen Zhu and Zheng Zhang and Zhexu Wang and Zhilin Yang and Zhiqi Huang and Zihao Huang and Ziyao Xu and Zonghan Yang and Zongyu Lin},
      year={2025},
      eprint={2501.12599},
      archivePrefix={arXiv},
      primaryClass={cs.AI},
      url={https://arxiv.org/abs/2501.12599}, 
}

@misc{qwen2025qwen25technicalreport,
      title={Qwen2.5 Technical Report}, 
      author={Qwen and : and An Yang and Baosong Yang and Beichen Zhang and Binyuan Hui and Bo Zheng and Bowen Yu and Chengyuan Li and Dayiheng Liu and Fei Huang and Haoran Wei and Huan Lin and Jian Yang and Jianhong Tu and Jianwei Zhang and Jianxin Yang and Jiaxi Yang and Jingren Zhou and Junyang Lin and Kai Dang and Keming Lu and Keqin Bao and Kexin Yang and Le Yu and Mei Li and Mingfeng Xue and Pei Zhang and Qin Zhu and Rui Men and Runji Lin and Tianhao Li and Tianyi Tang and Tingyu Xia and Xingzhang Ren and Xuancheng Ren and Yang Fan and Yang Su and Yichang Zhang and Yu Wan and Yuqiong Liu and Zeyu Cui and Zhenru Zhang and Zihan Qiu},
      year={2025},
      eprint={2412.15115},
      archivePrefix={arXiv},
      primaryClass={cs.CL},
      url={https://arxiv.org/abs/2412.15115}, 
}

@misc{zhao2025absolutezeroreinforcedselfplay,
      title={Absolute Zero: Reinforced Self-play Reasoning with Zero Data}, 
      author={Andrew Zhao and Yiran Wu and Yang Yue and Tong Wu and Quentin Xu and Yang Yue and Matthieu Lin and Shenzhi Wang and Qingyun Wu and Zilong Zheng and Gao Huang},
      year={2025},
      eprint={2505.03335},
      archivePrefix={arXiv},
      primaryClass={cs.LG},
      url={https://arxiv.org/abs/2505.03335}, 
}

@misc{gao2024omnimathuniversalolympiadlevel,
      title={Omni-MATH: A Universal Olympiad Level Mathematic Benchmark For Large Language Models}, 
      author={Bofei Gao and Feifan Song and Zhe Yang and Zefan Cai and Yibo Miao and Qingxiu Dong and Lei Li and Chenghao Ma and Liang Chen and Runxin Xu and Zhengyang Tang and Benyou Wang and Daoguang Zan and Shanghaoran Quan and Ge Zhang and Lei Sha and Yichang Zhang and Xuancheng Ren and Tianyu Liu and Baobao Chang},
      year={2024},
      eprint={2410.07985},
      archivePrefix={arXiv},
      primaryClass={cs.CL},
      url={https://arxiv.org/abs/2410.07985}, 
}

@article{Slow_Thinking_with_LLMs_3_Preview,
  title={STILL-3-1.5B-preview: Enhancing Slow Thinking Abilities of Small Models through Reinforcement Learning
},
  author={RUCAIBox STILL Team},
  url={https://github.com/RUCAIBox/Slow_Thinking_with_LLMs},
  year={2025}
}
\bibliographystyle{tmlr}

\clearpage
\appendix
\section{Appendix}

\subsection{\textcolor{black}{Training Dynamics: \adarft vs. Fixed Curriculum vs. PPO}}
\label{sec:training_dynamics_fc}

\begin{figure*}[ht]
    \centering
    \includegraphics[width=\textwidth]{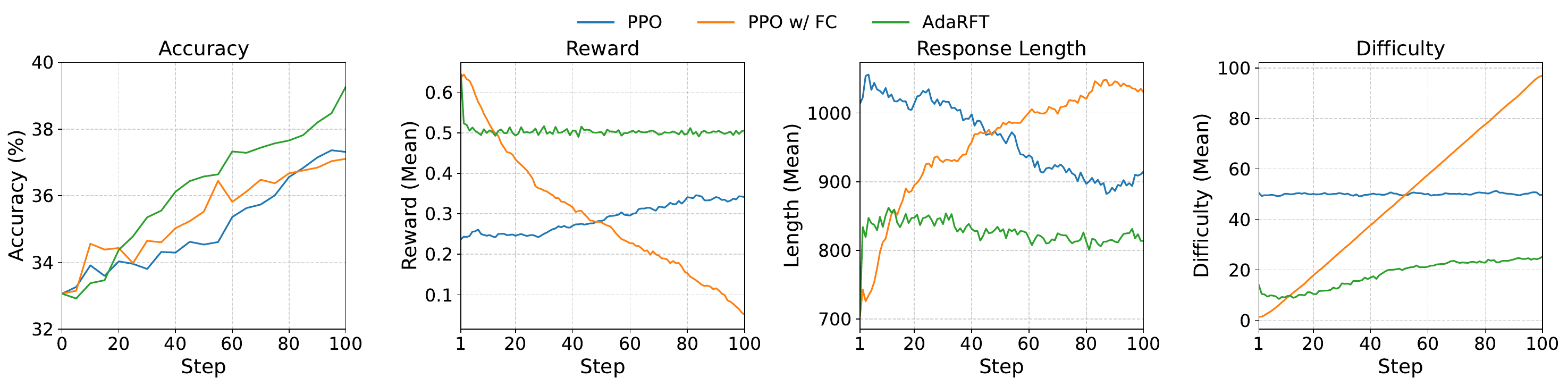}
    \caption{\textcolor{black}{Training dynamics of standard PPO, PPO with a fixed curriculum (PPO w/ FC), and \adarft on the Qwen 2.5 Math 1.5B model under the uniform data distribution. We plot accuracy, reward, response length, and average difficulty of sampled training problems over training steps. Curves are exponentially smoothed ($\alpha = 0.3$) for clarity.}}
    \label{fig:fc_result}
\end{figure*}

\textcolor{black}{To further contextualize the role of adaptive curricula, we analyze the training dynamics of fixed curriculum (PPO w/ FC), standard PPO, and \adarft when training the Qwen 2.5 Math 1.5B model on the uniform distribution. Figure~\ref{fig:fc_result} illustrates the evolution of accuracy, reward, response length, and sampled difficulty across the first 100 training steps.}

\textcolor{black}{The accuracy curves highlight the central tradeoff of fixed curricula. PPO w/ FC initially converges slightly faster than standard PPO, benefiting from early exposure to easier problems. However, it ultimately underperforms compared to \adarft because its difficulty schedule increases independently of the model’s actual learning progress. As a result, the model is pushed into harder problem regimes too quickly. This misalignment is clearly visible in the reward curve. Early in training, PPO w/ FC achieves rewards well above 0.5, indicating that the model is initially exposed to problems that are easier than its current capability. However, as training progresses, the fixed curriculum increases difficulty at a rate that outpaces the model’s learning speed. As a result, the reward drops below 0.5 and continues declining, showing that the model is increasingly confronted with problems it cannot yet solve. This mismatch limits the effectiveness of the updates and ultimately leads to slower convergence compared to an adaptive curriculum.}

\textcolor{black}{The response length patterns reinforce this interpretation. Because longer responses typically correspond to more difficult reasoning tasks, the rapid increase in response length under PPO w/ FC shows that the curriculum escalates difficulty faster than the model can adapt. In contrast, standard PPO maintains more stable lengths but lacks the structured progression necessary for efficient learning. \adarft, by comparison, keeps response lengths moderate and gradually increasing, consistent with its difficulty traces: the curriculum raises problem difficulty only when the model’s reward stays near the target value, ensuring that the model always receives examples that are challenging but solvable.}

\textcolor{black}{This particular dynamic suggest a general underlying issue: since the model’s perceived difficulty changes over the course of training, a fixed curriculum cannot remain aligned, and different model-dataset combinations may experience extended periods of being over-challenged or under-challenged. This mismatch limits the effectiveness of updates and leads to slower convergence compared to an adaptive curriculum. In contrast, \adarft is designed to handle precisely this challenge: by adjusting the difficulty schedule based on the model’s reward signal, it continually matches training difficulty to the model’s evolving capability, ensuring sustained learning progress throughout training. Without the ability to adjust to real-time model performance, fixed schedules risk either overwhelming the model or wasting compute on overly easy tasks. \adarft avoids both extremes by maintaining the average reward near 0.5, automatically pacing the introduction of harder problems as the model improves. This adaptive alignment between task difficulty and model capability leads to smoother reward trajectories, controlled response lengths, and ultimately more stable and efficient learning.}

\subsection{Detailed Breakdown of Model Performance}
Table~\ref{tab:result_step100} provides a per-benchmark breakdown of final accuracy at step 100 across all models, data setups, and training methods. Results are reported on six diverse mathematical reasoning benchmarks spanning grade-school arithmetic (GSM8K), competition-style problems (MATH 500, AMC 23, AIME 24), and advanced proof-oriented datasets (OlympiadBench, Minerva Math). Across both model scales and all data distributions, \adarft either matches or outperforms PPO-based baselines on the majority of individual benchmarks, resulting in consistent gains in the overall average accuracy.

For Qwen 2.5 Math 1.5B, improvements from \adarft are particularly pronounced on GSM8K, MATH 500, AMC 23, and AIME 24 under skew-difficult and uniform settings, indicating stronger generalization across both structured and competition-style problems. Even in the skew-easy setting, where baseline PPO already performs well, \adarft maintains or slightly improves performance without introducing regressions on harder benchmarks such as OlympiadBench and Minerva Math. For the larger Qwen 2.5 7B model, gains are more modest but remain consistent, with \adarft achieving higher average accuracy across all three data distributions and stable performance on the most challenging benchmarks. Overall, these results confirm that the efficiency gains of \adarft do not come at the cost of final model quality, and in practice often translate into improved performance across a broad range of mathematical reasoning tasks.

\label{sec:model_performance_details}
\begin{table*}[t]
\centering
\resizebox{\textwidth}{!}{%
\begin{tabular}{cllccccccc}
\toprule
Model & Setup & Method &
\begin{tabular}{c}GSM8K\end{tabular} &
\begin{tabular}{c}MATH\\500\end{tabular} &
\begin{tabular}{c}Olympiad\\Bench\end{tabular} &
\begin{tabular}{c}Minerva\\Math\end{tabular} &
\begin{tabular}{c}AMC 23\\ (Avg@8)\end{tabular} &
\begin{tabular}{c}AIME 24\\ (Avg@8)\end{tabular} &
Average \\
\midrule
\multirow{9}{*}{\begin{tabular}{c}Qwen 2.5\\Math\\1.5B\end{tabular}} 
 & \multirow{3}{*}{skew-difficult} 
   & \adarft & \textbf{74.00} & \textbf{66.40} & \textbf{20.36} & \textbf{15.07} & \textbf{55.00} & \textbf{12.08} & \textbf{40.48} \\
 &   & PPO & 69.67 & 64.60 & 20.65 & 12.87 & 47.50 & 9.17 & 37.41 \\
 &   & PPO (w/ Filter) & 71.65 & 62.40 & 20.06 & \textbf{15.07} & 45.00 & 9.17 & 37.22 \\
 &   & \textcolor{black}{PPO (w/ FC)} & \textcolor{black}{72.55} & \textcolor{black}{66.40} & \textcolor{black}{20.95} & \textcolor{black}{14.34} & \textcolor{black}{45.00} & \textcolor{black}{4.06} & \textcolor{black}{37.22} \\
\cmidrule(lr){2-10}
 & \multirow{3}{*}{uniform} 
   & \adarft & \textbf{74.53} & \textbf{66.20} & \textbf{21.99} & 14.34 & \textbf{57.50} & \textbf{12.08} & \textbf{41.11} \\
 &   & PPO & 71.95 & 65.20 & 21.10 & \textbf{15.81} & 42.50 & 6.67 & 37.20 \\
 &   & PPO (w/ Filter) & 72.63 & 65.80 & 20.21 & 13.60 & 45.00 & 10.00 & 37.87 \\
 &   & \textcolor{black}{PPO (w/ FC)} & \textcolor{black}{71.95} & \textcolor{black}{65.60} & \textcolor{black}{19.61} & \textcolor{black}{14.71} & \textcolor{black}{42.50} & \textcolor{black}{9.27} & \textcolor{black}{37.27} \\
\cmidrule(lr){2-10}
 & \multirow{3}{*}{skew-easy} 
   & \adarft & 73.62 & 66.20 & 19.91 & \textbf{13.97} & \textbf{55.00} & 9.17 & \textbf{39.18} \\
 &   & PPO & 72.71 & \textbf{67.40} & 19.17 & \textbf{13.97} & 45.00 & \textbf{12.50} & 38.46 \\
 &   & PPO (w/ Filter) & \textbf{74.75} & 65.20 & \textbf{20.36} & 13.60 & 45.00 & 10.00 & 38.15 \\
 &   & \textcolor{black}{PPO (w/ FC)} & \textcolor{black}{72.40} & \textcolor{black}{66.20} & \textcolor{black}{20.06} & \textcolor{black}{13.24} & \textcolor{black}{50.00} & \textcolor{black}{6.67} & \textcolor{black}{38.09} \\
\cmidrule(lr){1-10}
\multirow{9}{*}{\begin{tabular}{c}Qwen 2.5\\7B\end{tabular}} 
 & \multirow{3}{*}{skew-difficult} 
   & \adarft & \textbf{90.98} & 71.40 & \textbf{25.85} & 22.43 & \textbf{52.50} & \textbf{15.83} & \textbf{46.83} \\
 &   & PPO & 89.69 & 71.20 & 23.33 & 23.53 & 50.00 & 11.25 & 44.17 \\
 &   & PPO (w/ Filter) & 88.48 & \textbf{72.20} & 24.37 & 25.00 & 50.00 & 12.08 & 45.35 \\
 &   & \textcolor{black}{PPO (w/ FC)} & \textcolor{black}{89.92} & \textcolor{black}{72.00} & \textcolor{black}{24.52} & \textcolor{black}{25.37} & \textcolor{black}{47.50} & \textcolor{black}{13.96} & \textcolor{black}{45.54} \\
\cmidrule(lr){2-10}
 & \multirow{3}{*}{uniform} 
   & \adarft & \textbf{90.14} & 72.60 & \textbf{24.96} & 24.26 & \textbf{55.00} & 14.58 & \textbf{46.92} \\
 &   & PPO & 89.31 & 72.40 & 23.63 & \textbf{25.37} & 42.50 & \textbf{15.00} & 44.70 \\
 &   & PPO (w/ Filter) & 89.08 & \textbf{74.40} & 23.18 & 22.43 & 45.00 & 13.33 & 44.57 \\
 &   & \textcolor{black}{PPO (w/ FC)} & \textcolor{black}{89.84} & \textcolor{black}{72.40} & \textcolor{black}{23.92} & \textcolor{black}{23.90} & \textcolor{black}{42.50} & \textcolor{black}{13.33} & \textcolor{black}{44.32} \\
\cmidrule(lr){2-10}
 & \multirow{3}{*}{skew-easy} 
   & \adarft & \textbf{90.14} & 72.60 & \textbf{25.56} & 23.16 & \textbf{50.00} & \textbf{14.17} & \textbf{45.94} \\
 &   & PPO & 89.39 & \textbf{73.60} & 23.33 & \textbf{24.26} & 47.50 & 13.33 & 45.07 \\
 &   & PPO (w/ Filter) & 89.31 & 71.60 & 24.22 & 23.90 & 47.50 & 13.33 & 44.98 \\
 &   & \textcolor{black}{PPO (w/ FC)} & \textcolor{black}{89.46} & \textcolor{black}{72.20} & \textcolor{black}{24.37} & \textcolor{black}{23.90} & \textcolor{black}{37.50} & \textcolor{black}{13.33} & \textcolor{black}{43.46} \\
\bottomrule
\end{tabular}%
}
\caption{Accuracy (\%) at step 100 for every model, setup, and benchmark. \adarft in this table refers to \adarft instantiated with PPO, i.e., \adarft(PPO).}
\label{tab:result_step100}
\end{table*}

\section{Implementation Details}
\label{sec:implementation_details}

\subsection{Training Configuration}
We trained both the actor and critic models using the PPO algorithm on a single node with 8 A100 GPUs. Each model was trained for approximately 100 optimization steps using the \texttt{veRL} library \citep{sheng2024hybridflow}. We used two model variants: \texttt{Qwen2.5-7B} and \texttt{Qwen2.5-MATH-1.5B}. The latter has a shorted context window, so we adjusted the max response length and the sequence parallel size accordingly.

Table~\ref{tab:full_hyperparam_comparison} summarizes the core hyperparameter settings used across all three algorithms: PPO, GRPO, and REINFORCE++. We highlight both shared defaults and algorithm-specific overrides, including KL treatment modes, rollout settings, and critic configurations.

\subsection{\textcolor{black}{Deriving the Target-Difficulty Update Rule from a Linear Mapping}}
\label{sec:derive_target_update_rule}

\textcolor{black}{
A central component of our curriculum mechanism is the update of the target difficulty
\(T\)
based on the model's observed reward performance.
While the final update rule (Eq.~\ref{eq:difficulty_update}) involves a hyperbolic tangent,
it is in fact a smooth and stabilized variant of a standard linear mapping between
reward space and difficulty space.
We derive it here for clarity.
}

\paragraph{\textcolor{black}{Step 1: Linear mapping between two intervals.}}
\textcolor{black}{
The classical linear interpolation formula for mapping a value
\(v \in [x, y]\) to a target interval \([a, b]\) is
}
\begin{equation}
    \textcolor{black}{
    v' = a + \frac{(v - x)(b - a)}{y - x}.
    }
    \label{eq:linear_map}
\end{equation}

\textcolor{black}{
If we directly map the average reward
\(R_{\mathrm{avg}} \in [r_{\min}, r_{\max}]\)
to the difficulty range
\([d_{\min}, d_{\max}]\),
we obtain
}
\begin{equation}
\textcolor{black}{
\begin{aligned}
& T_{\mathrm{naive}}(R_{\mathrm{avg}}) = d_{\min} + \frac{(R_{\mathrm{avg}} - r_{\min})(d_{\max} - d_{\min})}
              {r_{\max} - r_{\min}} .
\end{aligned}
}
\end{equation}

\textcolor{black}{
In our main setting,
\(r_{\min} = 0\), \(r_{\max} = 1\),
and \([d_{\min}, d_{\max}] = [0, 100]\),
so the naive mapping simplifies to
}
\begin{equation}
    \textcolor{black}{
    T_{\mathrm{naive}}(R_{\mathrm{avg}}) = 100\, R_{\mathrm{avg}}.
    }
\end{equation}

\paragraph{\textcolor{black}{Step 2: Mapping reward deviation instead of absolute reward.}}
\textcolor{black}{
For curriculum learning, we do not wish to reassign a new difficulty level at every step.
Instead, we aim to \emph{adjust} the current target difficulty
depending on whether the model is performing above or below a desired target success rate
\(\beta\).
We therefore consider the deviation
}
\begin{equation}
    \textcolor{black}{
    \delta = R_{\mathrm{avg}} - \beta.
    }
\end{equation}

\textcolor{black}{
Given \(R_{\mathrm{avg}} \in [0, 1]\),
the deviation satisfies
\(
\delta \in [r_{\min} - \beta,\; r_{\max} - \beta]
= [-\beta,\; 1 - \beta].
\)
With the common choice \(\beta = 0.5\), this becomes
\(
\delta \in [-0.5, 0.5].
\)
}

\textcolor{black}{
Applying the linear mapping rule \eqref{eq:linear_map}
from the deviation range \([-0.5,\, 0.5]\)
to a symmetric difficulty-change interval \([-\Delta, \Delta]\)
yields
}
\begin{equation}
\textcolor{black}{
\begin{aligned}
\Delta T_{\mathrm{lin}}(\delta)
&= -\Delta
  + \frac{(\delta - (-0.5))(\Delta - (-\Delta))}{0.5 - (-0.5)} \\
&= 2\Delta\,\delta .
\end{aligned}
}
\end{equation}

\textcolor{black}{
Thus the naive linear controller becomes
}
\begin{equation}
    \textcolor{black}{
    T'_{\mathrm{lin}}
    = T + 2\Delta\, (R_{\mathrm{avg}} - \beta).
    }
\end{equation}

\textcolor{black}{
This already captures the desired behavior:
difficulty increases when performance exceeds the target,
decreases when performance falls short,
and remains stable when \(R_{\mathrm{avg}} = \beta\).
}

\paragraph{\textcolor{black}{Step 3: Stabilizing the update via a smooth saturating nonlinearity.}}
\textcolor{black}{
A purely linear controller may cause excessively large changes
when the reward deviation is large or noisy.
To obtain a stable update rule,
we replace the linear term with a smooth,
odd, saturating nonlinearity.
The hyperbolic tangent is a natural choice:
it behaves linearly near zero
(which recovers the linear mapping)
and saturates as its argument grows.
}

\textcolor{black}{
We therefore define a smoothed difficulty adjustment
}
\begin{equation}
    \textcolor{black}{
    \Delta T(\delta)
    = \eta \cdot \tanh\bigl(\alpha (R_{\mathrm{avg}} - \beta)\bigr).
    }
\end{equation}

\textcolor{black}{
Here,
\(\eta\) sets the maximum update magnitude
and
\(\alpha\) controls the sensitivity around the target reward.
For small deviations,
\(\tanh(z) \approx z\),
so locally
}
\begin{equation}
    \textcolor{black}{
    \Delta T(\delta) \approx \eta \alpha (R_{\mathrm{avg}} - \beta),
    }
\end{equation}

\textcolor{black}{
recovering a linear controller with effective slope \(\eta \alpha\)
while ensuring global boundedness.
}

\paragraph{\textcolor{black}{Step 4: Clipping to the valid difficulty range.}}
\textcolor{black}{
To ensure the target difficulty remains within the observed range of the data,
we apply a final clipping:
}
\begin{equation}
\label{eq:difficulty_update}
    \textcolor{black}{
    T'
    =
    \mathrm{clip}\!\left(
        T + \eta \cdot \tanh\bigl(\alpha (R_{\mathrm{avg}} - \beta)\bigr),
        \; d_{\min},\; d_{\max}
    \right).
    }
\end{equation}

\textcolor{black}{
The full update rule \eqref{eq:difficulty_update} is therefore a direct,
smoothed generalization of a naive linear mapping
between reward deviations and difficulty adjustments.
It preserves the intuitive behavior of the linear controller near the target reward,
while the saturating nonlinearity and clipping ensure stable,
bounded, and data-consistent curriculum updates.
}

\textcolor{black}{
Because the reward is bounded in $[0, 1]$ and the difficulty metric spans $[0, 100]$,
we set the step size $\eta = 50$ to align their scales.
The modulation parameter $\alpha = 2$ ensures smooth and controlled progression throughout training.
}

\begin{table*}[h]
\centering
\resizebox{\textwidth}{!}{
\begin{tabular}{llccc}
\toprule
\textbf{Category} & \textbf{Parameter} & \textbf{PPO} & \textbf{GRPO} & \textbf{REINFORCE++} \\
\midrule
\multicolumn{5}{l}{\textit{Algorithm-Specific Settings}} \\
\midrule
\multirow{6}{*}{General} 
  & Advantage estimator              & GAE & GRPO & REINFORCE++ \\
  & Gamma ($\gamma$)                & 1.0 & --- & --- \\
  & Lambda ($\lambda$)              & 1.0 & --- & --- \\
  & Batch size                      & 1024 & 1024 & 1024 \\
  & Max prompt length               & 1024 & 1024 & 1024 \\
  & Gradient checkpointing          & Enabled & Enabled & Enabled \\

\midrule
\multirow{10}{*}{Actor} 
  & Learning rate                   & $1 \times 10^{-6}$ & $1 \times 10^{-6}$ & $1 \times 10^{-6}$ \\
  & Mini-batch size                 & 1024 & 1024 & 1024 \\
  & Dynamic batch size              & Enabled & Enabled & Enabled \\
  & KL penalty role                    & Reward & Loss & Loss \\
  & KL loss type                    & Fixed & Low-variance KL & MSE \\
  & KL loss coefficient ($\beta$)   & 0.001 & 0.001 & 0.001 \\
  & Entropy coefficient             & 0.001 & 0.001 & 0 \\
  & Clip ratio                      & 0.2 & 0.2 & 0.2 \\
  & Gradient clipping               & 1.0 & 1.0 & 1.0 \\
  & Sequence parallel size          & Model-specific & Model-specific & Model-specific \\

\midrule
\multirow{6}{*}{Rollout} 
  & Backend                         & vLLM & vLLM & vLLM \\
  & Tensor model parallel size      & 2 & 2 & 2 \\
  & Rollouts per sample             & 1 & 8 & 1 \\
  & Nucleus sampling $p$                         & 1.0 & 1.0 & 1.0 \\
  & GPU memory utilization          & 0.5 & 0.5 & 0.5 \\
  & Sampling temperature            & 1.0 & 1.0 & 1.0 \\

\midrule
\multirow{3}{*}{Critic} 
  & Warmup steps                    & 0 & --- & --- \\
  & Learning rate                   & $1 \times 10^{-5}$ & --- & --- \\
  & Sequence parallel size          & Model-specific & --- & --- \\

\midrule
\multicolumn{5}{l}{\textit{Model-Specific Overrides (shared across all algorithms)}} \\
\midrule
\multirow{3}{*}{Qwen2.5-7B} 
  & Max response length             & 8000 & 8000 & 8000 \\
  & Sequence parallel size          & 2 & 2 & 2 \\
  & Max token length / GPU          & 8000 & 8000 & 8000 \\

\multirow{3}{*}{Qwen2.5-MATH-1.5B} 
  & Max response length             & 3000 & 3000 & 3000 \\
  & Sequence parallel size          & 1 & 1 & 1 \\
  & Max token length / GPU          & 16000 & 16000 & 16000 \\

\midrule
\multicolumn{5}{l}{\textit{\adarft Parameters}} \\
\midrule
\multirow{5}{*}{Curriculum Learning} 
  & Target reward ($\beta$)         & 0.5 & 0.5 & 0.5 \\
  & Sensitivity ($\alpha$)          & 2 & 2 & 2 \\
  & Step size ($\eta$)              & 50 & 50 & 50 \\
  & Initial difficulty ($T$)        & 0 & 0 & 0 \\
\bottomrule
\end{tabular}
}
\caption{Comparison of training hyperparameters for PPO, GRPO, and REINFORCE++ using the veRL library. Shared defaults are used unless overridden.}
\label{tab:full_hyperparam_comparison}
\end{table*}

\subsection{Prompt for Difficulty Estimation Using LLM as a Judge}
\label{sec:difficulty-estimation-prompt}
The prompt used for difficulty estimation (as described in Section~\ref{sec:difficulty_estimation}) is shown in Table~\ref{tab:difficulty1}, Table~\ref{tab:difficulty2}, and Table~\ref{tab:difficulty3}. The descriptions of the difficulty scales and examples are sourced from the AoPS Wiki.\footnote{\url{https://artofproblemsolving.com/wiki/index.php/AoPS_Wiki:Competition_ratings}} Although GPT-4o is prompted to rate problem difficulty on a scale from 1 to 10, we found that over 95\% of the problems fall within the range of 1 to 5. Therefore, we clip the scores and use a revised scale from 1 to 5. In addition to integer scores, we also allow half-point increments such as 1.5, 2.0, and 2.5 for finer-grained difficulty estimation.

\begin{table*}[t]
\begin{tcolorbox} [colback=yellow!10]
{\bf Prompt for Difficulty Estimation (Part 1)} 

\tcblower

\textbf{\# Math Problem}\\
\{problem\}\\

\textbf{\# Your Task}\\
You are a subject matter expert in mathematics tasked with evaluating the difficulty level of individual math problems. Your assessment should be objective and based on a detailed difficulty scale provided below.
Your judgment will help calibrate and categorize problems for use in educational settings or assessments. Be thorough, fair, and consistent in your evaluation.\\

\textbf{\# Difficulty Scale}\\
1: Problems strictly for beginner, on the easiest elementary school or middle school levels (MOEMS, MATHCOUNTS School, AMC 8 1-10, AMC 10 1-10, easier AMC 12 1-5, and others that involve standard techniques introduced up to the middle school level), most traditional middle/high school word problems.

1.5: Problems for stronger beginner students, on the level of the middling problems in most middle school contests (AMC 8 11-20, harder AMC 10 1-10, AMC 12 1-5, and those others that force students to apply their school-level knowledge to slightly more challenging problems), traditional middle/high school word problems with more complex problem solving.

2: For motivated beginners, harder questions from the previous categories (AMC 8 21-25, MATHCOUNTS Chapter (Sprint 21-30, Target 6-8), MATHCOUNTS States/Nationals, AMC 10 11-15, AMC 12 5-10, easiest AIME 1-3)

2.5: More advanced beginner problems, hardest questions from previous categories (Harder AMC 8 21-25, harder MATHCOUNTS States questions, AMC 10 16-20, AMC 12 11-15, usual AIME 1-3)

3: Early intermediate problems that require more creative thinking (harder MATHCOUNTS National questions, AMC 10 21-25, AMC 12 15-20, hardest AIME 1-3, usual AIME 4-6).

4: Intermediate-level problems (AMC 12 21-25, hardest AIME 4-6, usual AIME 7-10).

5: More difficult AIME problems (11-13), simple proof-based Olympiad-style problems (early JBMO questions, easiest USAJMO 1/4).

6: High-leveled AIME-styled questions (14/15). Introductory-leveled Olympiad-level questions (harder USAJMO 1/4 and easier USAJMO 2/5, easier USAMO and IMO 1/4).

7: Tougher Olympiad-level questions, may require more technical knowledge (harder USAJMO 2/5 and most USAJMO 3/6, extremely hard USAMO and IMO 1/4, easy-medium USAMO and IMO 2/5).

8: High-level Olympiad-level questions (medium-hard USAMO and IMO 2/5, easiest USAMO and IMO 3/6).

9: Expert Olympiad-level questions (average USAMO and IMO 3/6).

9.5: The hardest problems appearing on Olympiads which the strongest students could reasonably solve (hard USAMO and IMO 3/6).

10: Historically hard problems, generally unsuitable for very hard competitions (such as the IMO) due to being exceedingly tedious, long, and difficult (e.g. very few students are capable of solving on a worldwide basis).\\

\end{tcolorbox}
\caption{\label{tab:difficulty1}Prompt for difficulty estimation using LLM as a judge.}
\end{table*}

\begin{table*}[t]
\begin{tcolorbox} [colback=yellow!10]
{\bf Prompt for Difficulty Estimation (Part 2)} 

\tcblower
\textbf{\# Examples}\\
For reference, here are some sample problems from each of the difficulty levels 1-10:

<1: Jamie counted the number of edges of a cube, Jimmy counted the numbers of corners, and Judy counted the number of faces. They then added the three numbers. What was the resulting sum? (2003 AMC 8, Problem 1)

1: How many integer values of $x$ satisfy $|x| < 3\pi$? (2021 Spring AMC 10B, Problem 1)

1.5: A number is called flippy if its digits alternate between two distinct digits. For example, $2020$ and $37373$ are flippy, but $3883$ and $123123$ are not. How many five-digit flippy numbers are divisible by $15?$ (2020 AMC 8, Problem 19)

2: A fair $6$-sided die is repeatedly rolled until an odd number appears. What is the probability that every even number appears at least once before the first occurrence of an odd number? (2021 Spring AMC 10B, Problem 18)

2.5: $A$, $B$, $C$ are three piles of rocks. The mean weight of the rocks in $A$ is $40$ pounds, the mean weight of the rocks in $B$ is $50$ pounds, the mean weight of the rocks in the combined piles $A$ and $B$ is $43$ pounds, and the mean weight of the rocks in the combined piles $A$ and $C$ is $44$ pounds. What is the greatest possible integer value for the mean in pounds of the rocks in the combined piles $B$ and $C$? (2013 AMC 12A, Problem 16)

3: Triangle $ABC$ with $AB=50$ and $AC=10$ has area $120$. Let $D$ be the midpoint of $\overline{AB}$, and let $E$ be the midpoint of $\overline{AC}$. The angle bisector of $\angle BAC$ intersects $\overline{DE}$ and $\overline{BC}$ at $F$ and $G$, respectively. What is the area of quadrilateral $FDBG$? (2018 AMC 10A, Problem 24)

3.5: Find the number of integer values of $k$ in the closed interval $[-500,500]$ for which the equation $\log(kx)=2\log(x+2)$ has exactly one real solution. (2017 AIME II, Problem 7)

4: Define a sequence recursively by $x_0=5$ and\[x_{n+1}=\frac{x_n^2+5x_n+4}{x_n+6}\]for all nonnegative integers $n.$ Let $m$ be the least positive integer such that\[x_m\leq 4+\frac{1}{2^{20}}.\]In which of the following intervals does $m$ lie?

$\textbf{(A) } [9,26] \qquad\textbf{(B) } [27,80] \qquad\textbf{(C) } [81,242]\qquad\textbf{(D) } [243,728] \qquad\textbf{(E) } [729,\infty)$
(2019 AMC 10B, Problem 24 and 2019 AMC 12B, Problem 22)

\end{tcolorbox}
\caption{\label{tab:difficulty2}Prompt for difficulty estimation using LLM as a judge.}
\end{table*}

\begin{table*}[t]
\begin{tcolorbox} [colback=yellow!10]
{\bf Prompt for Difficulty Estimation (Part 3)} 
\tcblower
4.5: Find, with proof, all positive integers $n$ for which $2^n + 12^n + 2011^n$ is a perfect square. (USAJMO 2011/1)

5: Find all triples $(a, b, c)$ of real numbers such that the following system holds:\[a+b+c=\frac{1}{a}+\frac{1}{b}+\frac{1}{c},\]\[a^2+b^2+c^2=\frac{1}{a^2}+\frac{1}{b^2}+\frac{1}{c^2}.\](JBMO 2020/1)

5.5: Triangle $ABC$ has $\angle BAC = 60^{\circ}$, $\angle CBA \leq 90^{\circ}$, $BC=1$, and $AC \geq AB$. Let $H$, $I$, and $O$ be the orthocenter, incenter, and circumcenter of $\triangle ABC$, respectively. Assume that the area of pentagon $BCOIH$ is the maximum possible. What is $\angle CBA$? (2011 AMC 12A, Problem 25)

6: Let $\triangle ABC$ be an acute triangle with circumcircle $\omega,$ and let $H$ be the intersection of the altitudes of $\triangle ABC.$ Suppose the tangent to the circumcircle of $\triangle HBC$ at $H$ intersects $\omega$ at points $X$ and $Y$ with $HA=3,HX=2,$ and $HY=6.$ The area of $\triangle ABC$ can be written in the form $m\sqrt{n},$ where $m$ and $n$ are positive integers, and $n$ is not divisible by the square of any prime. Find $m+n.$ (2020 AIME I, Problem 15)

6.5: Rectangles $BCC_1B_2,$ $CAA_1C_2,$ and $ABB_1A_2$ are erected outside an acute triangle $ABC.$ Suppose that\[\angle BC_1C+\angle CA_1A+\angle AB_1B=180^{\circ}.\]Prove that lines $B_1C_2,$ $C_1A_2,$ and $A_1B_2$ are concurrent. (USAMO 2021/1, USAJMO 2021/2)

7: We say that a finite set $\mathcal{S}$ in the plane is balanced if, for any two different points $A$, $B$ in $\mathcal{S}$, there is a point $C$ in $\mathcal{S}$ such that $AC=BC$. We say that $\mathcal{S}$ is centre-free if for any three points $A$, $B$, $C$ in $\mathcal{S}$, there is no point $P$ in $\mathcal{S}$ such that $PA=PB=PC$.

Show that for all integers $n\geq 3$, there exists a balanced set consisting of $n$ points.
Determine all integers $n\geq 3$ for which there exists a balanced centre-free set consisting of $n$ points.
(IMO 2015/1)

7.5: Let $\mathbb{Z}$ be the set of integers. Find all functions $f : \mathbb{Z} \rightarrow \mathbb{Z}$ such that\[xf(2f(y)-x)+y^2f(2x-f(y))=\frac{f(x)^2}{x}+f(yf(y))\]for all $x, y \in \mathbb{Z}$ with $x \neq 0$. (USAMO 2014/2)

\end{tcolorbox}
\caption{\label{tab:difficulty3}Prompt for difficulty estimation using LLM as a judge.}
\end{table*}

\begin{table*}[t]
\begin{tcolorbox} [colback=yellow!10]
{\bf Prompt for Difficulty Estimation (Part 4)} 
\tcblower

8: For each positive integer $n$, the Bank of Cape Town issues coins of denomination $\frac1n$. Given a finite collection of such coins (of not necessarily different denominations) with total value at most most $99+\frac{1}{2}$, prove that it is possible to split this collection into $100$ or fewer groups, such that each group has total value at most $1$. (IMO 2014/5)

8.5: Let $I$ be the incentre of acute triangle $ABC$ with $AB\neq AC$. The incircle $\omega$ of $ABC$ is tangent to sides $BC, CA$, and $AB$ at $D, E,$ and $F$, respectively. The line through $D$ perpendicular to $EF$ meets $\omega$ at $R$. Line $AR$ meets $\omega$ again at $P$. The circumcircles of triangle $PCE$ and $PBF$ meet again at $Q$.

Prove that lines $DI$ and $PQ$ meet on the line through $A$ perpendicular to $AI$. (IMO 2019/6)

9: Let $k$ be a positive integer and let $S$ be a finite set of odd prime numbers. Prove that there is at most one way (up to rotation and reflection) to place the elements of $S$ around the circle such that the product of any two neighbors is of the form $x^2+x+k$ for some positive integer $x$. (IMO 2022/3)

9.5: An anti-Pascal triangle is an equilateral triangular array of numbers such that, except for the numbers in the bottom row, each number is the absolute value of the difference of the two numbers immediately below it. For example, the following is an anti-Pascal triangle with four rows which contains every integer from $1$ to $10$.\[\begin{array}{ c@{\hspace{4pt}}c@{\hspace{4pt}} c@{\hspace{4pt}}c@{\hspace{2pt}}c@{\hspace{2pt}}c@{\hspace{4pt}}c } \vspace{4pt}  & & & 4 & & &  \\\vspace{4pt}  & & 2 & & 6 & &  \\\vspace{4pt}  & 5 & & 7 & & 1 & \\\vspace{4pt}  8 & & 3 & & 10 & & 9 \\\vspace{4pt} \end{array}\]Does there exist an anti-Pascal triangle with $2018$ rows which contains every integer from $1$ to $1 + 2 + 3 + \dots + 2018$? (IMO 2018/3)

10: Prove that there exists a positive constant $c$ such that the following statement is true: Consider an integer $n > 1$, and a set $\mathcal S$ of $n$ points in the plane such that the distance between any two different points in $\mathcal S$ is at least 1. It follows that there is a line $\ell$ separating $\mathcal S$ such that the distance from any point of $\mathcal S$ to $\ell$ is at least $cn^{-1/3}$.

(A line $\ell$ separates a set of points S if some segment joining two points in $\mathcal S$ crosses $\ell$.) (IMO 2020/6)\\

\textbf{\# Return format}\\
Please return the corresponding difficulty scale (integer) in \textbackslash box\{\}

\end{tcolorbox}
\caption{\label{tab:difficulty4}Prompt for difficulty estimation using LLM as a judge.}
\end{table*}

\end{document}